\documentclass[runningheads]{llncs}
\usepackage[T1]{fontenc}
\usepackage{graphicx}
\usepackage{booktabs}
\usepackage[misc]{ifsym}
\newcommand{\corr}{(\Letter)}
\usepackage{mwe}
\usepackage{comment}

\usepackage{subcaption}
\usepackage{comment}
\usepackage{siunitx}
\usepackage{tikz}
\usetikzlibrary{arrows.meta, positioning, decorations.pathreplacing}
\usepackage{float}
\usepackage{graphicx}
\usepackage{multicol}  
\usepackage{nccmath}
\usepackage{graphicx} 
\usepackage{amssymb}
\usepackage{url}
\usepackage{wrapfig}
\usepackage[hidelinks]{hyperref}
\usepackage[symbol]{footmisc}

\newcommand{\XXX}{{LCBM}}
\newcommand{\XXXdef}{{Learnable Concept-Based Model}}

\newcommand{\myparagraph}[1]{\smallskip\textbf{#1. }}
\newcommand{\rev}[1]{#1}

\begin{document}

\title{Towards Better Generalization and Interpretability in Unsupervised Concept-Based Models}

\titlerunning{Towards Better Generalization and Interpretability in Unsupervised CBMs}


\author{Francesco De Santis\inst{1} \corr \and
Philippe Bich\inst{1} \and
Gabriele Ciravegna\inst{1} \and
Pietro Barbiero\inst{2} \and
Danilo Giordano\inst{1} \and
Tania Cerquitelli\inst{1}}

\authorrunning{F. De Santis et al.}

\institute{Politecnico di Torino, Torino, 10129, Italy \email{\{name.surname\}@polito.it}
\and
Universita’ della Svizzera Italiana, Lugano, 6900, Switzerland. \email{pietro.barbiero@usi.ch}
}

\maketitle              

\begin{abstract}
    To increase the trustworthiness of deep neural networks, it is critical to improve the understanding of how they make decisions. This paper introduces a novel unsupervised concept-based model for image classification, named \XXXdef\ (\XXX) which models concepts as random variables within a Bernoulli latent space. Unlike traditional methods that either require extensive human supervision or suffer from limited scalability, our approach employs a reduced number of concepts without sacrificing performance. We demonstrate that \XXX{} surpasses existing unsupervised concept-based models in generalization capability and nearly matches the performance of black-box models. The proposed concept representation enhances information retention and aligns more closely with human understanding. A user study demonstrates the discovered concepts are also more intuitive for humans to interpret. Finally, despite the use of concept embeddings, we maintain model interpretability by means of a local linear combination of concepts\footnote[1]{Paper accepted at \textit{ECML-PKDD 2025.}}.
    
\keywords{CBM \and XAI \and Interpretable AI.}
\end{abstract}

\section{Introduction} 
Understanding the \textit{reason} why Deep Neural Networks (DNNs) make decisions is critical in today's society, as these models are increasingly deployed and affect people's lives. This concern has also led regulatory institutions to mandate interpretability and the possibility of challenging the decisions of deep neural networks as prerequisites for Artificial Intelligence (AI) 
systems~\cite{veale2021demystifying,panigutti2023role}. EXplainable AI (XAI) methods have emerged to address this challenge~\cite{ribeiro2016should,adadi2018peeking,guidotti2018survey}. However, several papers argue that feature importance explanations (such as saliency maps~\cite{zeiler2014visualizing,selvaraju2017grad}) have failed to achieve this goal, since showing where a network is looking is insufficient to explain the reasons behind its decisions~\cite{rudin2019stop,achtibat2023attribution}. To truly explain what the network has seen, many XAI methods are shifting toward explanations in terms of human-understandable attributes, or \textit{concepts}~\cite{kim2018interpretability,achtibat2023attribution,fel2023craft,poeta2023concept}. Concepts can be either extracted post-hoc~\cite{ghorbani2019towards} or directly inserted within the network representation to create a so-called concept-based model (CBMs, ~\cite{koh2020concept}).

\begin{wrapfigure}{r}{0.5\textwidth} 
    \centering
    \includegraphics[width=1\linewidth, trim=0 180 0 0, clip]{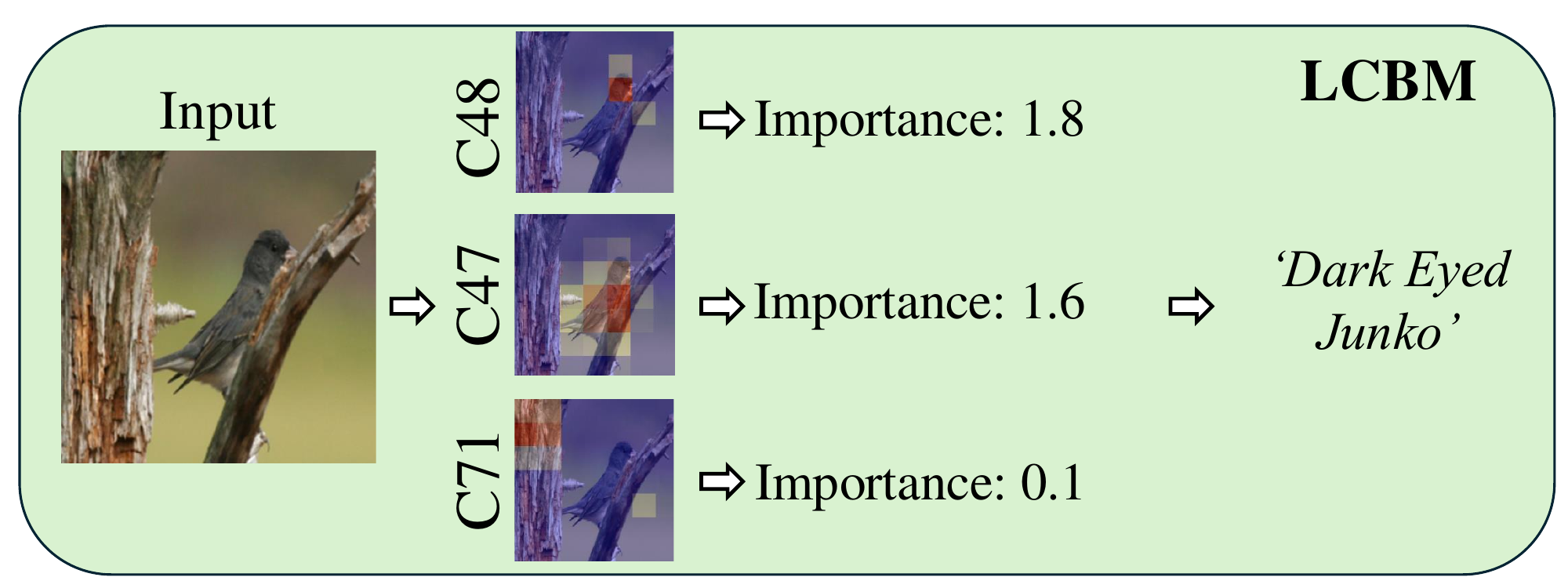} 
    \caption{\XXXdef\ (\XXX) learns a dictionary of unsupervised concepts. Unlike black-box models, \XXX\ classifies images interpretably using these concepts. Here, the image is correctly classified as \textit{Dark-eyed junco} by leveraging concepts C48 (eyes/beak), C47 (wings), and C71 (trunk/tree). Notably, C71, while present, is less relevant to bird species classification.} 
    \label{fig:visual_abs}
\end{wrapfigure}

CBMs can be created in a supervised way~\cite{koh2020concept,barbiero2022entropy,zarlenga2022concept} or through a dedicated unsupervised learning process~\cite{alvarez2018towards,chen2019looks,wang2023learning}. The latter approach enables the use of CBMs in contexts where concept annotations are unavailable and large language models (LLMs)~\cite{yang2023language,oikarinen2023label} lack sufficient knowledge. Yet, a fundamental challenge persists: standard unsupervised approaches rely on single-neuron activations to represent each concept, thereby limiting the amount of information that can be captured. This limitation creates a trade-off between interpretability and accuracy. The issue becomes even more pronounced when concise explanations are needed to avoid cognitive overload for users \cite{miller1956magical,lakkaraju2019faithful,ciravegna2023logic}. As demonstrated in our experiments, standard unsupervised approaches exhibit a significant performance gap compared to end-to-end methods in such scenarios, making unsuitable their effective deployment. In this paper, we demonstrate that by using unsupervised concept embeddings, we can create a highly effective \XXXdef{} (\XXX) employing a limited number of concepts. Our experiments show that this approach: i) overcomes the limited generalization of compared models, almost matching the performance of black-box models; ii) increases the representation capability of standard unsupervised concept layers in terms of information retention and alignment with human representation; iii) ensures that the extracted concepts are more interpretable, as highlighted by a user study; and iv) by providing the task prediction through a local linear combination of concepts, it retains task interpretability\footnote{Code to reproduce the proposed model is available at \href{https://github.com/francescoTheSantis/Unsupervised-Concept-Attention-Model/}{\textit{https://github.com/LCBM}}.}. 

\section{Related Work} 

Concept-based XAI (C-XAI) aims to provide human-understandable explanations by using concepts as intermediate representations \cite{kim2018interpretability,poeta2023concept,rudin2019stop}. While supervised approaches \cite{koh2020concept,zarlenga2022concept} rely on predefined symbols, unsupervised models autonomously extract concepts by modifying a network's internal representation through unsupervised learning, prototypical representations, or hybrid techniques \cite{alvarez2018towards,koh2020concept}.  

\myparagraph{Unsupervised Concept Basis}  
These methods learn disentangled representations in the model’s latent space by grouping samples based on fundamental characteristics. They typically achieve this via input reconstruction \cite{alvarez2018towards,wang2023learning} or unsupervised losses \cite{zhang2018interpretable,wang2023learning}. In \cite{zhang2018interpretable}, convolutional filters act as unsupervised concepts, maximizing mutual information between images and filter activations. SENN \cite{alvarez2018towards} employs an autoencoder to derive clustered representations and generate class-concept relevance scores. BotCL \cite{wang2023learning} enhances SENN with attention-based concept scoring and contrastive loss. Compared to these, \XXX{} introduces concept embeddings for richer representations, improving the generalization-interpretability trade-off.

\myparagraph{Prototype Concepts}  
This approach encodes training example traits as prototypes within the network, comparing them to input samples for prediction. \cite{li2018deep} explains predictions via prototype similarity, using an autoencoder for dimensionality reduction. ProtoPNet \cite{chen2019looks} extracts prototypes representing image subparts, while HPNet \cite{hase2019interpretable} organizes prototypes hierarchically for classification across taxonomy levels. Despite providing useful example-based explanations, these models constrain representation capacity. Our approach enhances performance by leveraging richer representations while retaining prototype-based interpretability, as shown in concept dictionaries.

\myparagraph{Hybrid Approaches}  
Recent research explores hybrid models that integrate supervised and unsupervised concepts \cite{marconato2022glancenets,sawada2022concept} or leverage pre-trained LLMs \cite{oikarinen2023label,yang2023language,yuksekgonul2022posthoc}. The variational approach in \cite{marconato2022glancenets} shares similarities with ours but relies on single neurons and partial supervision, limiting scalability. Methods leveraging LLMs assume sufficient knowledge for zero-shot concept annotations, yet this depends on the underlying model \cite{srivastava2024vlg}. For instance, CLIP \cite{radford2021learning}, despite its popularity, exhibits low concept accuracy even in contexts similar to its pre-training, as confirmed by our experiments.

\section{Methodology}
In an unsupervised concept-based setting, the objective is to make predictions using a set of abstract, human-interpretable concepts that are not predefined but must be directly inferred from the data. 
To address this challenge, we propose a set of desiderata that define the required properties of the learned concepts:


\begin{itemize} 
\item \textit{Representativity} \cite{bengio2013representation}: Concepts should capture key features of the input data.
\item \textit{Completeness} \cite{yeh2020completeness}: Concepts should support strong task generalization. 
\item \textit{Alignment} \cite{poeta2023concept}: Concepts should correspond to human-understandable constructs. \end{itemize}

If concepts do not accurately represent the input data, they cannot be trusted for either inference or explainability. As a result, \textit{Representativity}, a common requirement in representation learning \cite{bengio2013representation}, is a necessary but not sufficient condition for an interpretable unsupervised concept-based model: 
\textit{Completeness} is also essential to ensure the concepts are useful for making task-specific predictions. Ultimately, 
interpretability is achieved only with an \textit{Alignment} with a human-defined representation. While this property is always met in supervised CBMs, in unsupervised contexts it is a major challenge. 

\subsection{\XXXdef}
In a supervised learning context, a CBM is trained to approximate the joint distribution $p(x, c, y)$, where $x$, $c$, and $y$ correspond to realizations of the random variables $X$ (images), $C$ (concepts), and $Y$ (class labels), respectively. In contrast to the supervised setting, where these variables are fully observable during training, the unsupervised scenario lacks knowledge about $C$. Therefore, it is only through marginalizing over $C$ that we can account for the combined effect of all possible values of $C$ on the relationship between $X$ and $Y$: 
\begin{equation}
    p(x, y) = \int_c p(x, c, y) \, dc
\end{equation}

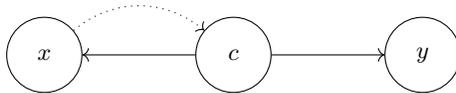
\begin{figure}[t]
    \centering
    \begin{tikzpicture}[
        node distance=1.5cm,
        roundnode/.style={circle, draw=black, minimum size=1cm, font=\small},
        ]
        
        \node[roundnode] (x) {$x$};
        \node[roundnode, right=of x] (c) {$c$};
        \node[roundnode, right=of c] (y) {$y$};
        
        \draw[->] (c) -- (x);
        \draw[->] (c) -- (y);
        
        \draw[dotted, ->, bend left=40] (x) to (c);

    \end{tikzpicture}
    \caption{Probabilistic Graphical Model. Solid arrows represent the data generating process. Dotted arrows represent inference.} 
    \label{fig:pgm}
\end{figure}

In order to address this problem, we introduce the \XXXdef\; (\XXX), a latent variable model enabling explanations and interventions in terms of a set of unsupervised concepts. Following~\cite{misino2022vael}, \XXX{} considers a data generating process in which concepts $C$ represent latent factors of variation for both $X$ and $Y$, as shown in the probabilistic graphical model in Figure~\ref{fig:pgm}.
Thus, the joint distribution factorizes as:
\begin{equation}
    p(x, y) = \int_c p(x, c, y) \mathrm{d}c = \int_c p(x \mid c) p(y \mid c) p(c) \mathrm{d}c
\end{equation}
where $p(y\mid c)$ is modelled as a categorical distribution parametrized by the task predictor $f$; $p(x \mid c)$ as a Gaussian distribution parametrized by the concept decoder $\psi$. Finally, $p(c)$ is a prior distribution over a set of unsupervised concepts.

During training, \XXX\; assumes to observe realizations of the random variables $X$ and $Y$ which hold new evidence we can use to update the prior $p(c)$. Since the computation of the true posterior $p(c \mid x, y)$ is intractable, \XXX\; amortizes inference needed for training by introducing an approximate posterior $q(c \mid x)$ parametrized by a neural network. Since at test time we can only observe $X$, we condition the approximate only on this variable.

\begin{figure*}[t]
    \centering
        \includegraphics[width=\textwidth]{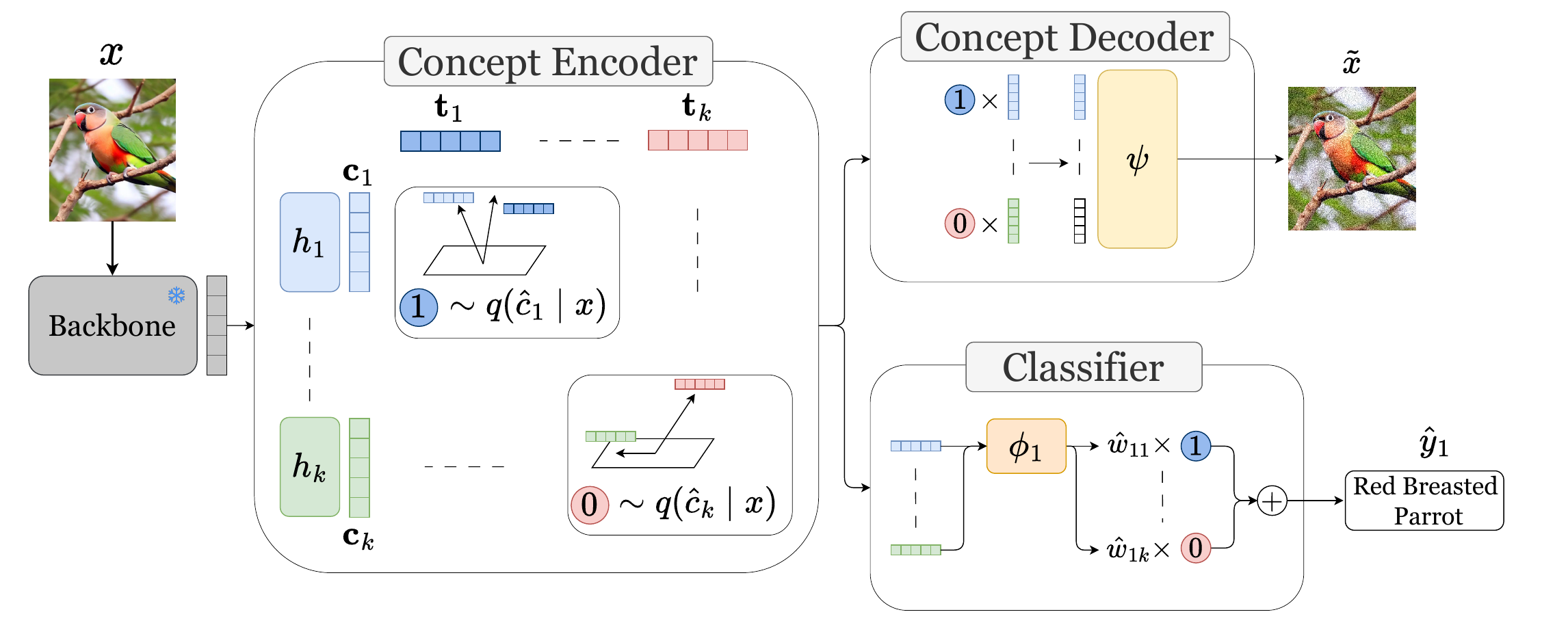} 
        \caption{\XXX{} schema. The concept encoder $q(c\mid x)$ provides the probability for each learnt concept $\hat{c}_j$ and the associated embeddings $\mathbf{c}_j$. Both concept scores and embeddings are used to predict the output class $p(y\mid c)$ and to reconstruct the input $p(x\mid c)$. } 
    \label{fig:schema}
\end{figure*}

\myparagraph{Optimization problem} \label{sec:optimisation}
\XXX{}s are trained to optimize the log-likelihood of tuples $(x, y)$. Following a variational inference approach, we optimize the evidence lower bound (ELBO) of the log-likelihood, which results as follows: 
\begin{equation}     \label{eq:elbo}
\footnotesize{\text{ELBO} = \overbrace{E_q[\text{log }p(x|c)]}^{\text{Representativity}}+\overbrace{E_q[\text{log }p(y|c)]}^{\text{Completeness}}-\overbrace{KL(q(c | x) \mid \mid  p(c))}^{\text{Alignment}}}
\end{equation}
For a complete derivation of the ELBO, see Appendix~\ref{app:derivation}. This likelihood has three components: a reconstruction term $p(x|c)$, whose maximization ensures concepts' \textit{Representativity}; a classifier $p(y\mid c)$, which quantifies concepts' \textit{Completeness}; and a Kullback–Leibler divergence term that encourages the approximate posterior $q(c \mid x)$ to remain close to a defined prior $p(c)$, promoting an \textit{Alignment} to human representations. 
To achieve all the desired properties, we must define a sufficiently rich concept representation. 
We describe the latter together with its prior in Section~\ref{sec:concept_representation},  the classifier in Section~\ref{sec:classifier} and the decoder in Section~\ref{sec:decoder}. For an overall visualization of \XXX, see Figure~\ref{fig:schema}.  

\subsection{Unsupervised Concept Representation}
\label{sec:concept_representation}

To model each concept in the concept representation $c$ in an unsupervised way, 
we define it as following a Bernoulli distribution. This choice 
reflects a discrete, binary nature of a ‘concept’ as an atomic unit of knowledge, inducing \textit{Alignment} and facilitating its comprehension and modification through human interventions. 
However, Bernoulli distributions may not be able to represent both the input and output distributions, ultimately creating a bottleneck in the representation of the model. 

To solve this issue, 
we associate each concept with a corresponding unsupervised concept embedding $\mathbf{c}_j \in \mathbf{C} \subseteq \mathbb{R}^d$. This embedding provides a richer representation of the concept, capturing further nuances (e.g., `color` and `size` if the concept represents a `vehicle`). 
The concept embedding $\mathbf{c}_j$ is derived  as a composition of two neural modules $ h \circ  g$. The latter is a frozen pre-trained backbone $g: X \rightarrow E$ mapping input $X$ into an intermediate embedding space $E \subseteq \mathbb{R}^s$, while the first module, $h_j: E \rightarrow \mathbf{C}$, is a per-concept MLP producing the concept embedding $\mathbf{c}_j = h_j(g(x))$. To ensure that this embedding is representative of the intended concept and allows interventions, we assign each concept a prototype $\mathbf{t}_j \in \mathbb{R}^d$, which serves as a learned reference within the embedding space of $\mathbf{c}_j$. To compute the concept score $\hat{c}_j$, we first calculate the alignment between the concept embedding $\mathbf{c}_j$ and its prototype $\mathbf{t}_j$ through their dot product $\mathbf{c}_j \cdot \mathbf{t}_j$, and transform it into a probability $\pi_j = \sigma(\mathbf{c}_j \cdot \mathbf{t}_j) \in [0,1]$ via a sigmoid function $\sigma$. Using $\pi_j$ as the probability parameter, we sample from a Bernoulli distribution  applying the reparametrization trick \cite{maddison2022concrete} to obtain the concept score $\hat{c}_j$. Thus, the final concept score is sampled from the following distribution:

\begin{equation} \hat{c}_j \sim q(\hat{c}_j\mid x) = \text{Bern}(\hat{c}_j ; \pi_j) \cdot p(\mathbf{c}_j \mid x), \end{equation}
where $p(\mathbf{c}_j\mid x)$ is the probability distribution parametrized by $h(g(x))$ and can be modelled either via Gaussian distributions \cite{kim2023probabilistic}, or through a degenerate Dirac delta distribution, without assuming any uncertainty. For the sake of simplicity, we choose the second approach.  

\myparagraph{Batch Prior Regularization}
\label{sec:prior} 
The parameter $\alpha$ parameterizes the Bernoulli prior in the KL term of Eq.~\ref{eq:elbo},  and determines the activation probability of each concept for every sample. It is fundamental to optimize KL divergence over a batch of sample rather than for each sample. Indeed, by setting $\alpha=0.2$, and performing the optimization for each sample, we force each concept to activate for each sample with $20\%$ confidence. This behaviour is far from optimal as we want \XXX{} to be sure about the presence or absence of a certain concept in a specific sample, i.e.,  producing $\pi_j\approx1$ for a sample which contains a specific concept and $\pi_j\approx0$ otherwise. To address this, we shift the KL  divergence optimization at the batch level by averaging the activation probabilities for a concept $j$ across the batch: $\bar{\pi}_j = \frac{1}{B}\sum_{z=1}^B \pi_{jz}$, where $B$ is the batch size. 

\subsection{Interpretable Classifier} 
\label{sec:classifier}
The classifier $f(\mathbf{c}, \hat{c})$ leverages both concept embeddings and concept scores to boost prediction accuracy without sacrificing interpretability. Specifically, each class prediction $\hat{y}_i$ is represented as a linear combination of concept scores $\hat{c}_j$ and associated weights  $\hat{w}_{ij} \in \mathbb{R}$,  where the weights are predicted over the concept embedding, i.e., $\hat{w}_{ij}=\phi_i(\mathbf{c}_{j})$, and $\phi_i$ is a class-specific function parameterized by a neural network. The output prediction $\hat{y}$ is then computed as $
\hat{y} = \underset{i}{\text{argmax }} p(y_i|c) = \underset{i}{\text{argmax }}\sum_j \hat{w}_{ij}\cdot\hat{c}_{ij}. $
Note that $\hat{w}_{ij}$ depends on the concept embedding $\mathbf{c}_j = g(x)$ predicted for a specific sample. This means that while the final prediction is provided by means of a linear classification over the concept scores, thus preserving locally the interpretability of standard CBMs, the network $\phi$ can predict different weights $\hat{w}_{mj}$ for different samples, thus overcoming the representation bottleneck of standard CBMs.

\subsection{Concept Decoder}
\label{sec:decoder}
As previously introduced, we parametrize the decoding function with a neural network $\psi$. To improve the image reconstruction capabilities, also in this case we rely on the concept embeddings $\mathbf{c}$. However, to still take into account the associated concept predictions $\hat{c}$,  we multiply the embeddings $\mathbf{c_j}$ by the corresponding concept prediction $\hat{c}_j$ before feeding them to the concept decoder. As a result,  the input is reconstructed as $\hat{x} = \psi(\hat{c}\cdot\mathbf{c})$.

\section{Experiments}

In this section, we present the experiments conducted to evaluate our proposed methodology. The experiments are designed to address the following key research questions: 

\begin{enumerate}
    \item \textbf{Generalization:} How effectively does the model generalize for classification tasks? Is the concept representation \textit{complete}?

    \item \textbf{Concept Representation Evaluation:} How much information is captured from $c$ with respect to both the input image $x$ and the label $y$? Are the learnt concepts \textit{representative} of the data? 

    \item \textbf{Concept Interpretability:} How interpretable are the learnt concepts produced? 
    Are they \textit{aligned} with human representations? 

    \item \textbf{Model Interpretability:} 
    Are the final predictions interpretable in terms of the discovered concepts? Can a user modify the concept prediction to extract counterfactual predictions? 
\end{enumerate}



\subsection{Experimental setting}
\label{sec:exp_settings}
In the following we report the dataset, metrics and baselines that we consider for evaluating and comparing our model.  We conducted experiments using two backbones $g$: ResNet-18 and ViT-base-patch32. The results for the ViT-base-patch32 backbone are included in Appendix \ref{app:app_vit}, as ProtoPNet is not designed for ViT-based backbones. Instead, we always use a decoder $\psi$  composed by five transposed convolutional layers. Additionally, an ablation regarding the concept embedding size is provided in Appendix~\ref{app:abl_emb}.

\myparagraph{Dataset} 
This study uses seven image classification datasets of varying complexity. We employ two MNIST~\cite{mnist} variants, \textit{Even/Odd} (digit parity) and \textit{Addition} (paired digits summed as labels). \textit{CIFAR-10} and \textit{CIFAR-100} contain 10 and 100 natural image classes, with models extracting 15 and 20 macro-class concepts, respectively~\cite{cifar10_100}. \textit{Tiny ImageNet} includes 200 classes but is tested on 30 concepts for added challenge~\cite{tiny_imagenet}. \textit{Skin Lesions} classifies dermatoscopic images into 4 macro categories~\cite{skinlesions}. Finally, \textit{CUB-200}~\cite{cub} covers 200 bird species with species and attribute annotations. Dataset details and preprocessing are in Appendix~\ref{app:app_dataset}.

\begin{figure*}[t]
    \centering
    \includegraphics[trim =0 0 20 0, clip, width=1\textwidth]{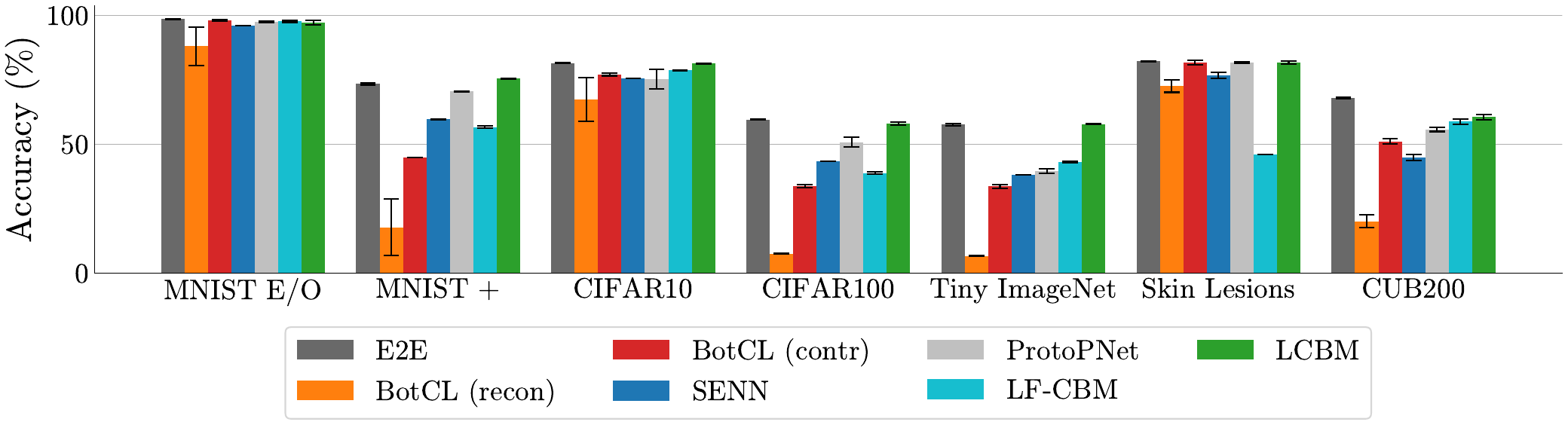}
    \caption{Comparison of the generalization performance across the evaluated datasets. \XXX{} consistently provides the highest generalization accuracy across concept-based models, closing the gap with end-to-end black box ones.}
    \label{fig:task_acc}
\end{figure*}

\myparagraph{Metrics}
We use specific metrics to address each research question. All results are reported with the mean and standard deviation, computed over the test sets by repeating the experiments with three different initialization seeds.

\begin{enumerate} 

\item \textbf{Generalization:} To assess the classification generalization performance, we compute the \textit{Task Accuracy}.

\item \textbf{Concept Representation Evaluation:} 
We employ the \textit{Information Plane} approach~\cite{info_plane} to analyze the information retained in the different concept representations. The information plane reports the evolution of the mutual information between the concept representation and both the input $x$ ($I(C,X)$) and the label $y$ ($I(C,Y)$) as the training epoch increases. 
For models reconstructing the input from the concepts, we assess the \textit{Input Reconstruction Error} by computing the Mean Squared Error (MSE) between the inputs $x$ and their reconstructions $\hat{x}$.

\item \textbf{Concept Interpretability:} 

For datasets with annotated concepts, we assess their alignment with the learnt concepts using the macro \textit{Concept F1 Score} (best-match approach) and the \textit{Concept Alignment Score (CAS)}~\cite{zarlenga2022concept} for concept representation alignment. Additionally, we conducted a user study with 72 participants, each answering 18 questions. The study evaluated \textit{Plausibility} by asking users to (i) select an image that best represents a given concept and (ii) identify an intruder image among those representing a single concept. It also assessed \textit{Human Understanding} by having participants assign a name to a set of images illustrating a concept. More details are available in Appendix~\ref{app:survey}. Finally, we provide qualitative insights through \textit{Concept Dictionaries}, showcasing images with the strongest activations for each concept.

\item \textbf{Model Interpretability:}  We perform \textit{Concept Interventions}~\cite{koh2020concept} to observe how model predictions change when concept predictions are modified. As positive concept interventions are non-trivial in unsupervised concept settings, we perform negative interventions~\cite{zarlenga2022concept}. Negative interventions involve randomly swapping the values of the concept scores with a given probability, expecting model accuracy to decrease as intervention probability increases. For \XXX, to switch a concept to inactive, we set $\hat{c}_j=0$, while to activate it, we set $\hat{c}_j$= 1 and replace the concept embedding with the concept prototype $\bar{c}_j = \mathbf{t_j}$. Additionally, we provide \textit{Qualitative Explanations} generated by \XXX{} using as concept importances the predicted weights multiplied by the concept predictions $w_{ij} \cdot \hat{c}_j$.

\end{enumerate}

\myparagraph{Baselines}
To compare the performance of the proposed approach, we test it against unsupervised approaches like \textit{SENN}~\cite{alvarez2018towards} and two variants of a SOTA model BotCL~\cite{wang2023learning}: \textit{BotCL (Recon)}, which employs an autoencoder-based approach to reconstruct the input image from the concept bottleneck, and \textit{BotCL (Contr)}, which applies a contrastive term to the loss to encourage distinct concept activations for different classes. Also, we consider a prototype based approach \textit{ProtoPNet}~\cite{chen2019looks} and Label-Free CBM (LF-CBM)~\cite{oikarinen2023label} a recent hybrid approach. If the concepts were known (e.g., MNIST Addition), we used CLIP to align the model with concept captions (e.g., this image contains the digit 4). If the concepts were unknown, we used the LLM to generate a list of possible concepts. Finally, we compare with a standard black-box model trained end-to-end (E2E).

\begin{figure*}[t]
    \centering
        \includegraphics[width=1\linewidth]{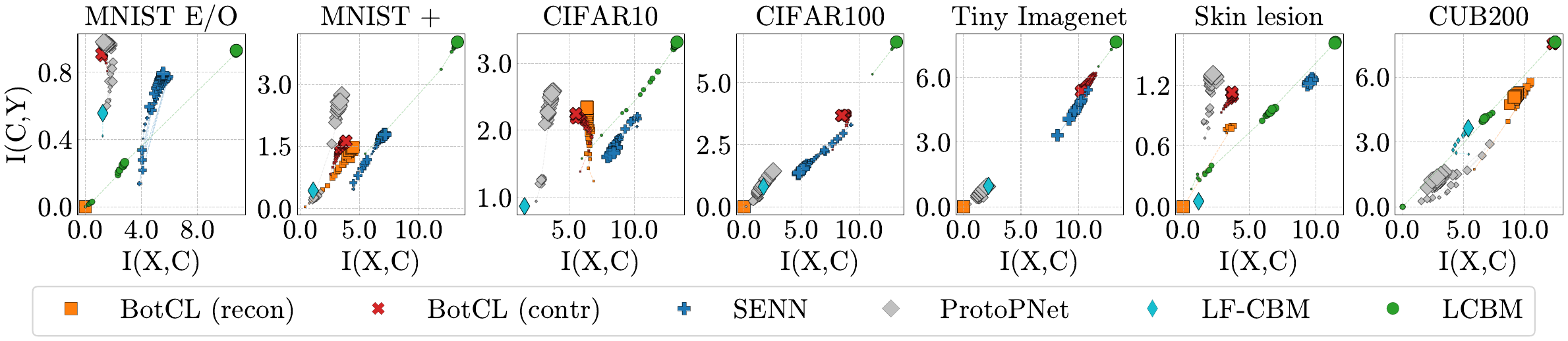} 
        \caption{Information Plane for the different models in terms of Mutual information between concept and input variables $I(X,C)$, and between the concept and output variables $I(C,Y)$.} 
    \label{fig:info_plane}
\end{figure*}

\begin{table*}[t]
    \centering
    \caption{We report the Input Reconstruction Error in terms of MSE for those methods that explicitly reconstruct the input. }
    \label{tab:mse}
    \resizebox{\linewidth}{!}{
    \begin{tabular}{@{}lccccccc@{}}
        \toprule
        & \textbf{MNIST E/O} & \textbf{MNIST Add.} & \textbf{CIFAR10} & \textbf{CIFAR100} & \textbf{Tiny ImageNet} & \textbf{Skin Lesion} & \textbf{CUB200} \\ \midrule
        BotCL           & 1.40 \footnotesize{± 0.09} & 1.57 \footnotesize{± 0.03} & 0.87 \footnotesize{± 0.02} & 0.82 \footnotesize{± 0.01} & 1.35 \footnotesize{± 0.06} & 0.72 \footnotesize{± 0.02} & 0.07 \footnotesize{± 0.01} \\
        SENN                    & 0.62 \footnotesize{± 0.02} & 0.93 \footnotesize{± 0.03} & 0.81 \footnotesize{± 0.04} & 0.74 \footnotesize{± 0.01} & 1.10 \footnotesize{± 0.02} & 0.61 \footnotesize{± 0.04} &  0.05 \footnotesize{±$\le$0.01} \\
        \textbf{\XXX{}} & \textbf{0.32 \footnotesize{±$\le$0.01}} & \textbf{0.71 \footnotesize{± 0.11}} & \textbf{0.51 \footnotesize{±$\le$0.01}} & \textbf{0.55 \footnotesize{±$\le$0.01}} & \textbf{0.72 \footnotesize{±$\le$0.01}} & \textbf{0.32 \footnotesize{±$\le$0.01}} & \textbf{0.05 \footnotesize{±$\le$0.01}}\\ 
        \bottomrule
    \end{tabular}
    }
\end{table*}

\subsection{Generalization} 
\label{subsec:generalization}
\myparagraph{\XXX{} is the most accurate interpretable model (Fig.~\ref{fig:task_acc})}
The proposed methodology significantly outperforms the baselines. In the most challenging scenario (Tiny ImageNet with only 30 concepts), it achieves up to a 50\% increase in task accuracy compared to the worst baseline (BotCL (recon)) and up to a 17\% improvement over the runner-up model, ProtoPNet. 
Our model consistently delivers the best generalization accuracy across all datasets, with higher gaps in challenging gaps with lower concept-class ratios. This result is valid even when we compare \XXX{} with LF-CBM which exploit the pre-existing knowledge within a VLM to extract concept annotations.
Only in the very simple MNIST Even/Odd dataset a few methods perform better, by a few decimals. 
We attribute this improvement to the unsupervised concept embeddings, which allows learning more \textit{complete} representations for task prediction.

\myparagraph{\XXX{} closes the gap with black-box models (Fig.~\ref{fig:task_acc})}
Figure~\ref{fig:task_acc} also shows that \XXX{} achieves results comparable to the E2E black-box model. The generalization loss is always less than 1-2\%. Notably, on the MNIST addition dataset, a setting where reasoning capabilities over concepts are required, our approach outperforms the black-box model with a task accuracy improvement of 2\%. Overall, \XXX{} demonstrates its capability to achieve high interpretability without sacrificing accuracy in unsupervised concept learning settings. For a more detailed representation of the results, check Table~\ref{tab:task_acc} in Appendix~\ref{app:task_acc}.

\subsection{Concept Representation Evaluation}

\myparagraph{\XXX{} concept representation retains more information regarding both the input and the output (Fig.~\ref{fig:info_plane})} 
The concept representation obtained through unsupervised concept embedding is significantly richer than that derived from simple concept scores.  
As training progresses, most baselines 
experience a reduction in mutual information with the input $I(X,C)$ while increasing the mutual information with the output $I(Y,C)$. This observation supports the conclusion that 
unsupervised CBM models tend to lose input-related information while attempting to optimize task performance~\cite{shwartz2017opening}, even for those models that explicitly require concepts to be representative of the input, such as SENN and BotCL (Recon). 
On the contrary, \XXX{} overcome this limitation by means of concept embeddings, which facilitate a better balance between competing objectives, as evidenced by the monotonic increase in mutual information with both the input and output during training.

\myparagraph{\XXX{} allows better input reconstruction (Tab.~\ref{tab:mse})}
To understand why the mutual information $I(X,C)$ of \XXX{} is consistently higher compared to other reconstruction-based unsupervised CBMs, we assess the Input Reconstruction Error in terms of MSE. As shown in Table~\ref{tab:mse}, \XXX{} achieves lower MSE in image reconstruction compared to the runner-up model (usually SENN), with values ranging from $0.18$ to $0.38$. We believe that concept embeddings facilitate more accurate and efficient reconstruction by allowing more information to flow to the decoder network when a concept is active ($\hat{c}_j=1$). Unlike other unsupervised models that can only pass a single value ($\hat{c}_j$), \XXX{} passes the entire associated concept embedding ($\bar{c}_j$) to the decoder.

\begin{table}[t]
    \centering
    \caption{Macro F1-score for candidate concepts with respect to existing human-representations for datasets on which the latter are available.}
    \resizebox{\linewidth}{!}{
    \begin{tabular}{@{}lccccc@{}}
        \toprule
        & \textbf{MNIST E/O} & \textbf{MNIST Add.} & \textbf{CIFAR100} & \textbf{Skin} & \textbf{CUB200}\\ \midrule
        BotCL (recon)                    & 0.47 \footnotesize{± 0.01} & 0.41 \footnotesize{± 0.01} & 0.38 \footnotesize{± 0.03} & 0.47 \footnotesize{± 0.02} & 0.34\footnotesize{± 0.01} \\
        BotCL (contr)                    & 0.47 \footnotesize{± 0.02} & 0.45 \footnotesize{± 0.02} & 0.40 \footnotesize{± 0.04} & 0.44 \footnotesize{± 0.03} & 0.37\footnotesize{± 0.02}\\
        SENN                             & 0.61 \footnotesize{± 0.02} & 0.58 \footnotesize{± 0.01} & 0.44 \footnotesize{± 0.02} & 0.52 \footnotesize{± 0.02} & 0.41\footnotesize{± 0.02}\\
        ProtoPNet                        & 0.26 \footnotesize{± 0.01} & 0.24 \footnotesize{± 0.01} & 0.31 \footnotesize{± 0.01} & 0.16 \footnotesize{± 0.02}& 0.28\footnotesize{± 0.03}\\
        LF-CBM                           &  0.52\footnotesize{± 0.01} & 0.50 \footnotesize{± 0.03}  & 0.45 \footnotesize{± 0.01} & 0.58 \footnotesize{± 0.01} & 0.45 \footnotesize{± 0.01} \\
        \midrule
        \textbf{\XXX{} (ours)}           & \textbf{0.88 \footnotesize{± 0.08}} & \textbf{0.81 \footnotesize{± 0.04}} & \textbf{0.60 \footnotesize{±$\le$0.01}} &  \textbf{0.58 \footnotesize{±$\le$0.01}}  &  \textbf{0.55 \footnotesize{±$\le$0.01}} \\ \bottomrule
    \end{tabular}
    }
    \label{tab:concepts_f1}
\end{table}

\begin{figure*}[t]
    \centering
        \includegraphics[width=\linewidth]{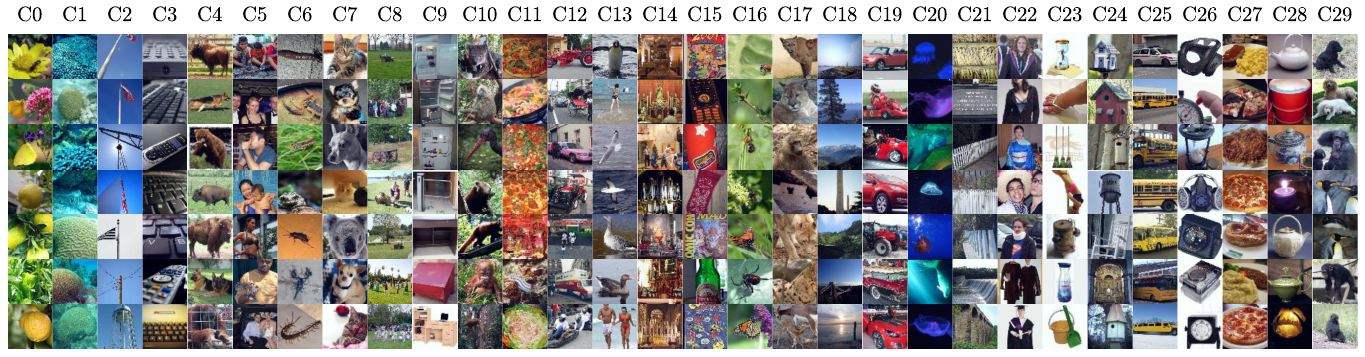} 
        \caption{\rev{Tiny-Imagenet dictionary produced by \XXX. Each column of images represents the set of 7 images that mostly activate each concept. Concept numbers are reported on top of each column.}} 
    \label{fig:imagenet_dict}
\end{figure*}


\subsection{Concept Interpretability} 
\label{subsec:interpret}
\myparagraph{\XXX{} concepts are more aligned to human-defined representations (Tab.~\ref{tab:concepts_f1})} 
For datasets with human-defined concept representations, we evaluate the alignment between these representations and those extracted by the compared concept learning methods. In Table~\ref{tab:concepts_f1}, we observe that \XXX{} learns concepts that are significantly more aligned with human-defined representations than existing methods. After matching the concept predictions with the concept annotations using the Hungarian algorithm~\cite{kuhn1955hungarian}, \XXX{} achieves an F1 score that is up to +0.36 higher than the runner-up, which is always LF-CBM. We remind, however, that this model has a huge intrinsic advantage, as the employed VLM is prompted to predict the concepts of each datasets. The fact that \XXX{} without any concept supervision achieves higher concept F1 scores than LF-CBM is impressive, but consistent with recent literature reporting poor LF-CBM concept accuracy \cite{srivastava2024vlg}.  
The improvement is also evident when considering the CAS, which evaluates the alignment using the embeddings instead of the concept scores. We report the CAS results in Table~\ref{tab:concepts_cas} in Appendix~\ref{app:cas_concept_align}.

\myparagraph{\XXX{} concepts are qualitatively distinguishable  (Fig.~\ref{fig:imagenet_dict})} 
While we quantitatively demonstrated that the \XXX{} concepts align with those of datasets equipped with annotations, for datasets lacking annotations, we examine the dictionaries representing the images that most strongly activate each concept, as proposed in~\cite{alvarez2018towards}. Fig.~\ref{fig:imagenet_dict} presents the dictionary generated by the model for the Tiny ImageNet dataset. Each column (concept) exhibits a recurring and distinguishable pattern. 
For example, concept \( C0 \) encompasses images of flowers and plants, whereas \( C2 \) appears to correspond to long, slender objects. Concept \( C4 \) includes large mammals such as bison and bears, while \( C7 \) represents close-up images of small animals. Additional details, including the discovered concept dictionaries for other datasets and the specifics of the dictionary generation process, can be found in Appendix~\ref{app:dictionary}.

\begin{figure}[t]
    \centering
    \begin{minipage}{0.48\textwidth}
        \centering
        \includegraphics[width=\linewidth]{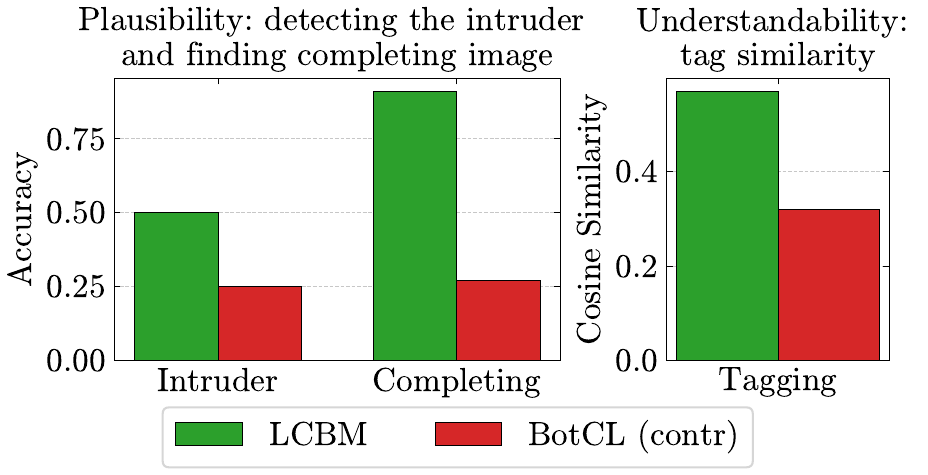}
        \caption{User study results. Left, user accuracy in detecting the intruder image and the image completing a set of images representing a concept. Right, the similarity of the tags employed by users to describe an extracted concept.}
        \label{fig:user_study}
    \end{minipage}
    \hfill
    \begin{minipage}{0.48\textwidth}
        \centering
        \includegraphics[width=\linewidth]{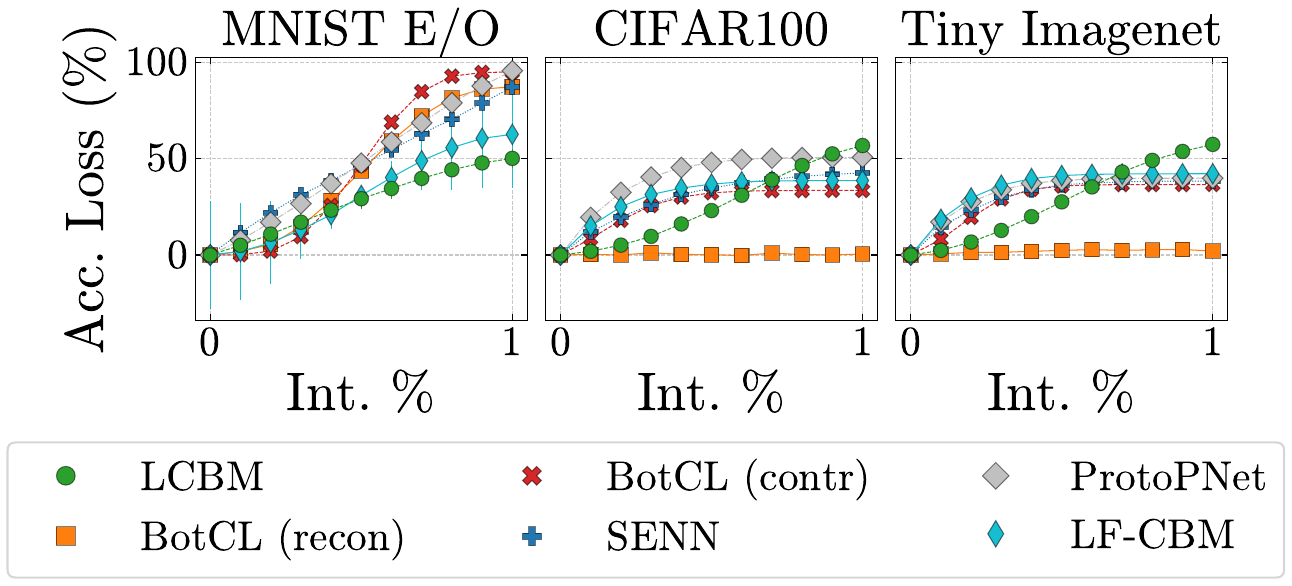}
        \caption{Negative interventions: percentage accuracy loss when increasing the intervention probability. The higher the accuracy loss, the higher the sensitivity of the model to human interventions. \rev{Results for all datasets can be found in Appendix~\ref{app:interventions}}.}
        \label{fig:intervention}
    \end{minipage}
\end{figure}

\myparagraph{\XXX{} concepts are more plausible and understandable to humans (Fig.\ref{fig:user_study})}
To quantitatively assess the quality of the representations, we conducted a user study comparing the plausibility and human-understandability of the concept extracted by our method and BotCL the SOTA baseline for unsupervised concept learning. 
Figure \ref{fig:user_study} shows that \XXX{} concepts enable users to find the intruder image and to complete the set of images much better than BotCL concepts, with an accuracy up to +25\% in terms of finding the right intruder image and up to +64\% in terms of selecting the completing image. Also, when assessing the understandability of the concepts we see a higher similarity up to +.35 of the embeddings of the terms employed to tag the concepts provided by \XXX{} than those of BotCL. The embeddings are generated using the multilingual sentence encoder “all-MiniLM-L6-v2”.  

\subsection{Model Interpretability}
\label{sec:model_intr}

\myparagraph{\XXX{} is sensitive to concept interventions (Fig.~\ref{fig:intervention})}
Figure~\ref{fig:intervention} shows that our methodology responds to interventions similarly to other baselines, except MNIST Even-Odd, where the precision drops only to $50\%$. In all other datasets, concept interventions are effective, with \XXX{} generally experiencing one of the highest accuracy losses when fully intervened, particularly on CIFAR100 and Tiny ImageNet. This is notable since embedding-based supervised CBMs typically resist interventions and require specialized training~\cite{espinosa2024learning}, whereas \XXX{} does not, suggesting its potential for more effective interventions in supervised settings.

\begin{figure}[t]
    \centering
    \begin{minipage}{0.49\textwidth}
        \centering
        \includegraphics[trim=0 0 10 0, clip, width=\textwidth]{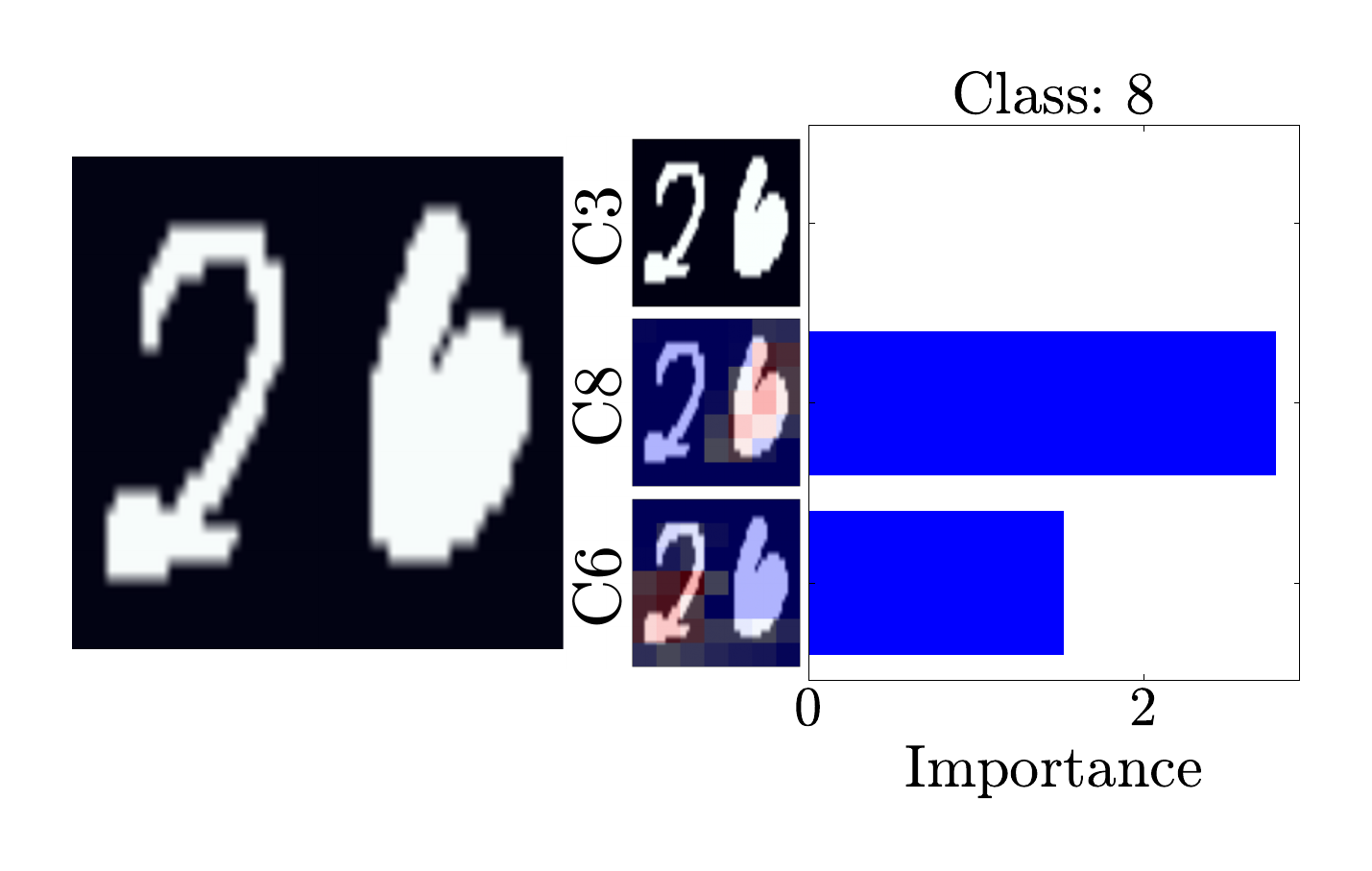} 
        \textbf{(a)} 
        \label{fig:explanation_8}
    \end{minipage}%
    \begin{minipage}{0.49\textwidth}
        \centering
        \includegraphics[trim=0 0 10 0, clip, width=\textwidth]{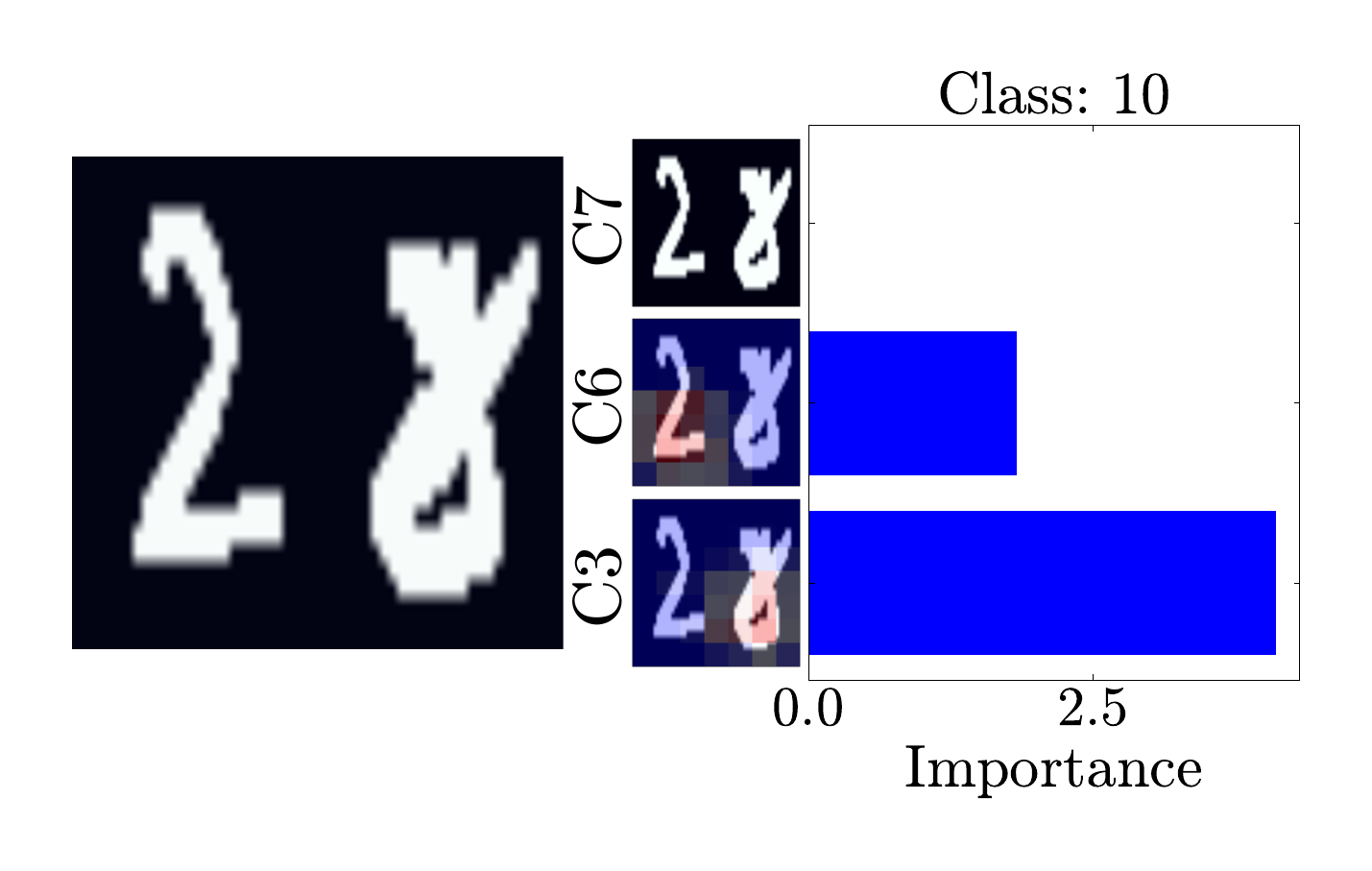} 
        \textbf{(b)} 
        \label{fig:explanation_10}
    \end{minipage} \\
    \begin{minipage}{0.49\textwidth}
        \centering
        \includegraphics[trim=0 0 10 0, clip, width=\textwidth]{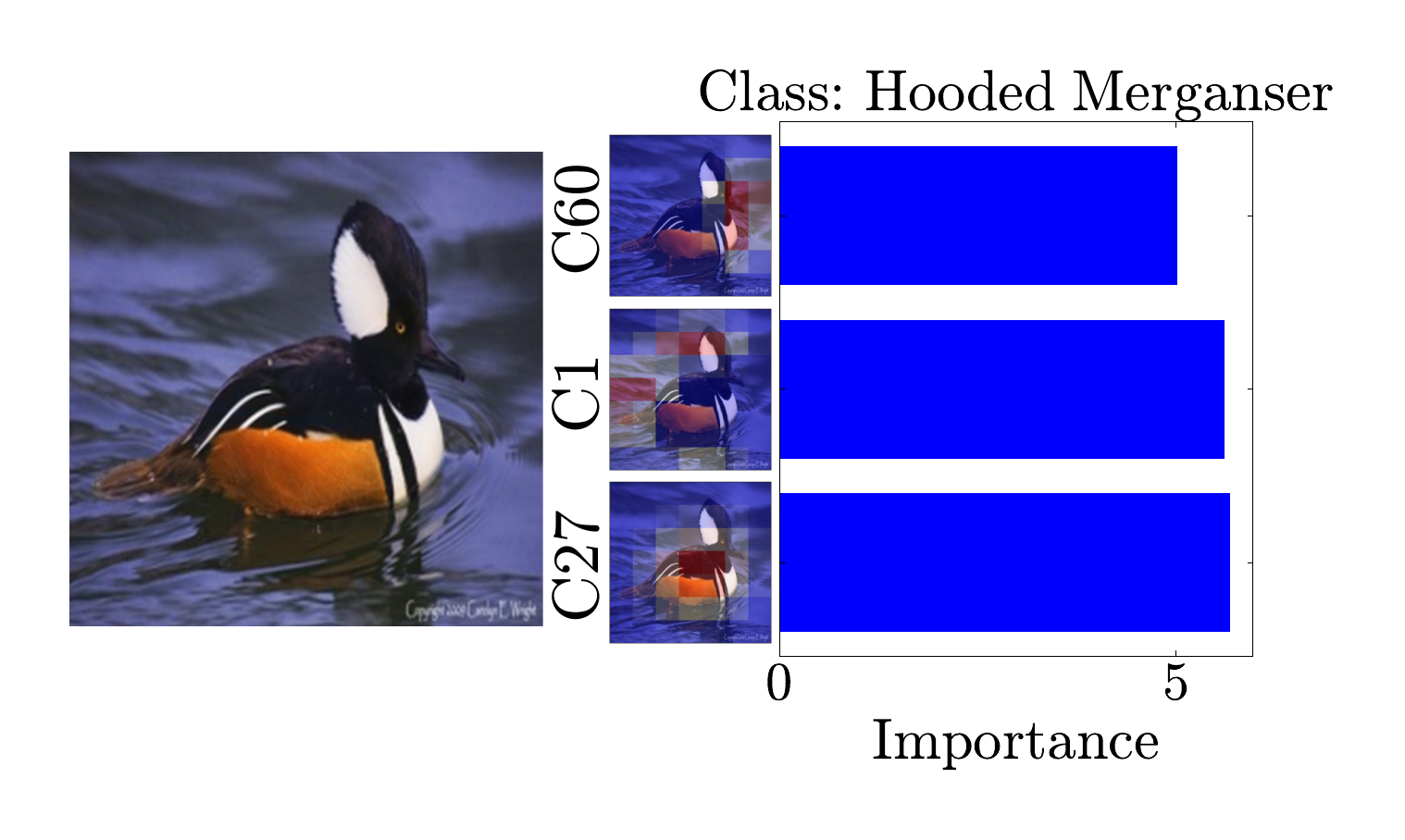} 
        \textbf{(c)} 
        \label{fig:hooded_merganser}
    \end{minipage}%
    \begin{minipage}{0.49\textwidth}
        \centering
        \includegraphics[trim=0 0 10 0, clip, width=\textwidth]{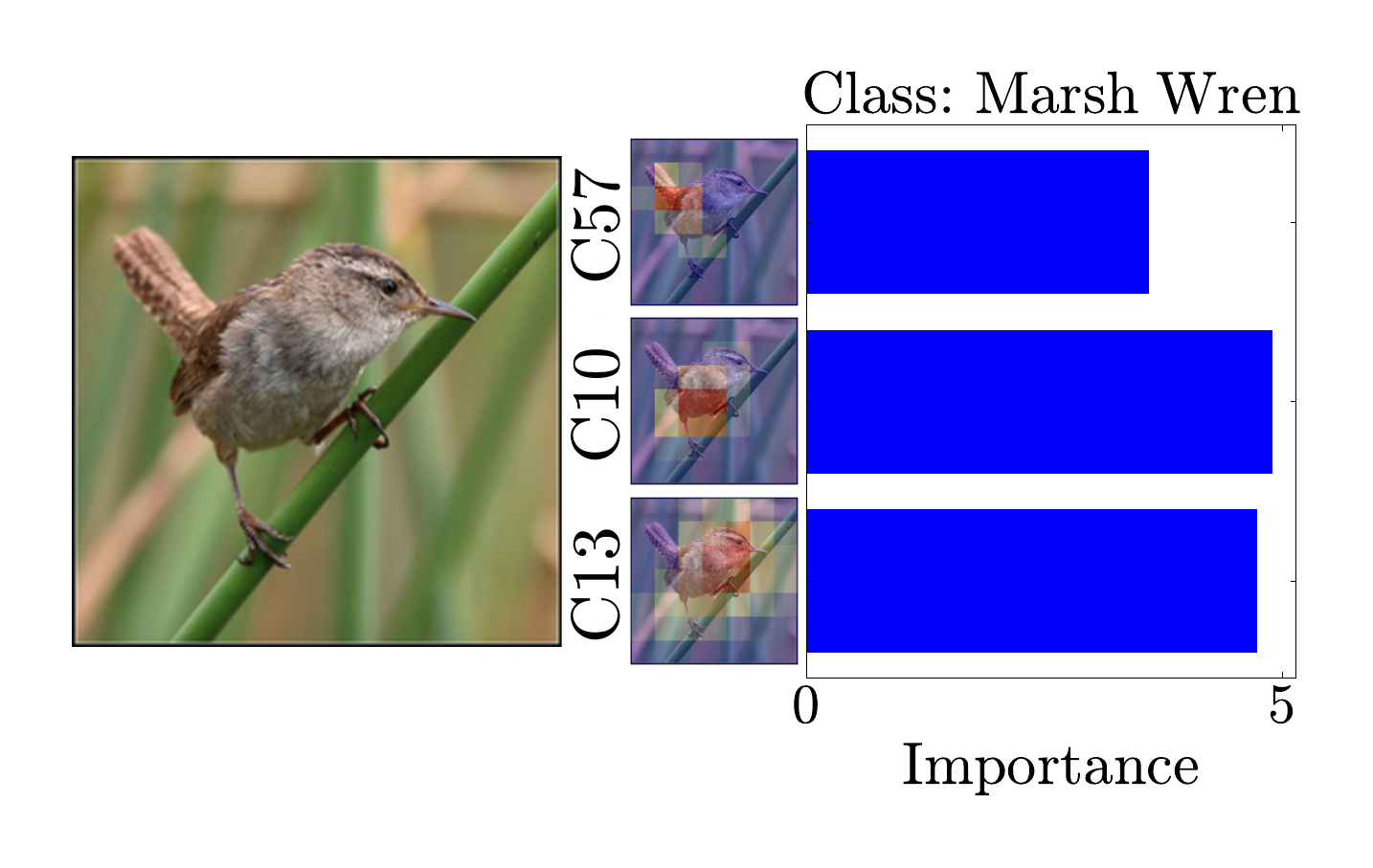} 
        \textbf{(d)} 
        \label{fig:marsh_wren}
    \end{minipage}
    \caption{Example of interpretable prediction on different datasets. We provide the concept importance together with the Grad-CAM for the most important concepts.}
    \label{fig:explanations}
\end{figure}

\myparagraph{\XXX{} provides interpretable predictions (Fig. \ref{fig:explanations})} 
Finally, we present sample explanations generated by \XXX. As illustrated in Figure~\ref{fig:explanations}.a, the image contains a "two" and a "six," with the model correctly predicting the sum as eight. The model identifies concepts $C6$ and $C8$ as important, which correspond to the learned concepts "two" and "six," as indicated by the concept dictionary (Figure~\ref{fig:mnist_sum_dict}) and further validated by the Grad-CAM results (shown on the y-axis), highlighting the respective digits in the image. Concept $C3$, which does not appear in the image, has an importance value of 0. Figure~\ref{fig:explanations}.c illustrates how an image of a bird is classified as a Hooded Merganser based on a triplet of concepts: $C27$ focuses on the orange wing, $C1$ on the crest extending from the back of the head, and $C60$ on the black beak. Additional sample explanations are provided in Appendix~\ref{app:explanations}.


\section{Conclusion}
This paper introduced a novel unsupervised concept learning model that leverages unsupervised concept embeddings. This approach enables improved generalization accuracy compared to traditional unsupervised Concept-Based Models (CBMs), while also enhancing the representation of concepts. Our experiments demonstrate that the extracted concepts better represent the input data and align more closely with human representations, as evidenced by the F1-score metric, CAS, and the findings from the user study.

\myparagraph{Limitations and Future work}
The first limitation of this work lies in the employed CNN decoder. While it helps extract meaningful unsupervised concept representations, it struggles to effectively decode the learned concepts. 
While our model reduces the human effort in understanding learned concepts, some manual inspection is still required. Vision Language Models (VLMs) could help fully automate concept labeling by using representative images, but this approach may be less effective in contexts where VLMs lack knowledge, which is the primary area of application for unsupervised CBMs. Finally, our experiments have been limited to image classification tasks. Extending the model to generative tasks presents a challenge and could be explored in future work.


%
%
%
\bibliographystyle{splncs04}
\bibliography{paper}

\begin{thebibliography}{10}
\providecommand{\url}[1]{\texttt{#1}}
\providecommand{\urlprefix}{URL }
\providecommand{\doi}[1]{https://doi.org/#1}

\bibitem{achtibat2023attribution}
Achtibat, R., Dreyer, M., Eisenbraun, I., Bosse, S., Wiegand, T., Samek, W., Lapuschkin, S.: From attribution maps to human-understandable explanations through concept relevance propagation. Nature Machine Intelligence  \textbf{5}(9),  1006--1019 (2023)

\bibitem{adadi2018peeking}
Adadi, A., Berrada, M.: Peeking inside the black-box: a survey on explainable artificial intelligence (xai). IEEE access  \textbf{6},  52138--52160 (2018)

\bibitem{alvarez2018towards}
Alvarez~Melis, D., Jaakkola, T.: Towards robust interpretability with self-explaining neural networks. Advances in neural information processing systems  \textbf{31} (2018)

\bibitem{barbiero2022entropy}
Barbiero, P., Ciravegna, G., Giannini, F., Li{\'o}, P., Gori, M., Melacci, S.: Entropy-based logic explanations of neural networks. In: Proceedings of the AAAI Conference on Artificial Intelligence. vol.~36, pp. 6046--6054 (2022)

\bibitem{bengio2013representation}
Bengio, Y., Courville, A., Vincent, P.: Representation learning: A review and new perspectives. IEEE transactions on pattern analysis and machine intelligence  \textbf{35}(8),  1798--1828 (2013)

\bibitem{chen2019looks}
Chen, C., Li, O., Tao, D., Barnett, A., Rudin, C., Su, J.K.: This looks like that: deep learning for interpretable image recognition. Advances in neural information processing systems  \textbf{32} (2019)

\bibitem{ciravegna2023logic}
Ciravegna, G., Barbiero, P., Giannini, F., Gori, M., Li{\'o}, P., Maggini, M., Melacci, S.: Logic explained networks. Artificial Intelligence  \textbf{314},  103822 (2023)

\bibitem{zarlenga2022concept}
Espinosa~Zarlenga, M., Barbiero, P., Ciravegna, G., Marra, G., Giannini, F., Diligenti, M., Shams, Z., Precioso, F., Melacci, S., Weller, A., Li\'{o}, P., Jamnik, M.: Concept embedding models: Beyond the accuracy-explainability trade-off. In: Koyejo, S., Mohamed, S., Agarwal, A., Belgrave, D., Cho, K., Oh, A. (eds.) Advances in Neural Information Processing Systems. vol.~35, pp. 21400--21413. Curran Associates, Inc. (2022)

\bibitem{espinosa2024learning}
Espinosa~Zarlenga, M., Collins, K., Dvijotham, K., Weller, A., Shams, Z., Jamnik, M.: Learning to receive help: Intervention-aware concept embedding models. Advances in Neural Information Processing Systems  \textbf{36} (2024)

\bibitem{fel2023craft}
Fel, T., Picard, A., Bethune, L., Boissin, T., Vigouroux, D., Colin, J., Cad{\`e}ne, R., Serre, T.: Craft: Concept recursive activation factorization for explainability. In: Proceedings of the IEEE/CVF Conference on Computer Vision and Pattern Recognition. pp. 2711--2721 (2023)

\bibitem{ghorbani2019towards}
Ghorbani, A., Wexler, J., Zou, J.Y., Kim, B.: Towards automatic concept-based explanations. Advances in neural information processing systems  \textbf{32} (2019)

\bibitem{guidotti2018survey}
Guidotti, R., Monreale, A., Ruggieri, S., Turini, F., Giannotti, F., Pedreschi, D.: A survey of methods for explaining black box models. ACM computing surveys (CSUR)  \textbf{51}(5),  1--42 (2018)

\bibitem{hase2019interpretable}
Hase, P., Chen, C., Li, O., Rudin, C.: Interpretable image recognition with hierarchical prototypes. In: Proceedings of the AAAI Conference on Human Computation and Crowdsourcing. vol.~7, pp. 32--40 (2019)

\bibitem{kim2018interpretability}
Kim, B., Wattenberg, M., Gilmer, J., Cai, C., Wexler, J., Viegas, F., et~al.: Interpretability beyond feature attribution: Quantitative testing with concept activation vectors (tcav). In: International conference on machine learning. pp. 2668--2677. PMLR (2018)

\bibitem{kim2023probabilistic}
Kim, E., Jung, D., Park, S., Kim, S., Yoon, S.: Probabilistic concept bottleneck models. In: Proceedings of the 40th International Conference on Machine Learning. ICML'23, JMLR.org (2023)

\bibitem{koh2020concept}
Koh, P.W., Nguyen, T., Tang, Y.S., Mussmann, S., Pierson, E., Kim, B., Liang, P.: Concept bottleneck models. In: International conference on machine learning. pp. 5338--5348. PMLR (2020)

\bibitem{cifar10_100}
Krizhevsky, A., et~al.: Learning multiple layers of features from tiny images  (2009)

\bibitem{kuhn1955hungarian}
Kuhn, H.W.: The hungarian method for the assignment problem. Naval research logistics quarterly  \textbf{2}(1-2),  83--97 (1955)

\bibitem{lakkaraju2019faithful}
Lakkaraju, H., Kamar, E., Caruana, R., Leskovec, J.: Faithful and customizable explanations of black box models. In: Proceedings of the 2019 AAAI/ACM Conference on AI, Ethics, and Society. pp. 131--138 (2019)

\bibitem{mnist}
Lecun, Y., Bottou, L., Bengio, Y., Haffner, P.: Gradient-based learning applied to document recognition. Proceedings of the IEEE  \textbf{86}(11),  2278--2324 (1998). \doi{10.1109/5.726791}

\bibitem{li2018deep}
Li, O., Liu, H., Chen, C., Rudin, C.: Deep learning for case-based reasoning through prototypes: A neural network that explains its predictions. In: Proceedings of the AAAI Conference on Artificial Intelligence. vol.~32 (2018)

\bibitem{maddison2022concrete}
Maddison, C.J., Mnih, A., Teh, Y.W.: The concrete distribution: A continuous relaxation of discrete random variables. In: International Conference on Learning Representations (2022)

\bibitem{marconato2022glancenets}
Marconato, E., Passerini, A., Teso, S.: Glancenets: Interpretable, leak-proof concept-based models. Advances in Neural Information Processing Systems  \textbf{35},  21212--21227 (2022)

\bibitem{miller1956magical}
Miller, G.A.: The magical number seven, plus or minus two: Some limits on our capacity for processing information. Psychological review  \textbf{63}(2), ~81 (1956)

\bibitem{misino2022vael}
Misino, E., Marra, G., Sansone, E.: Vael: Bridging variational autoencoders and probabilistic logic programming. Advances in Neural Information Processing Systems  \textbf{35},  4667--4679 (2022)

\bibitem{oikarinen2023label}
Oikarinen, T., Das, S., Nguyen, L.M., Weng, T.W.: Label-free concept bottleneck models. In: The Eleventh International Conference on Learning Representations (2023), \url{https://openreview.net/forum?id=FlCg47MNvBA}

\bibitem{panigutti2023role}
Panigutti, C., Hamon, R., Hupont, I., Fernandez~Llorca, D., Fano~Yela, D., Junklewitz, H., Scalzo, S., Mazzini, G., Sanchez, I., Soler~Garrido, J., et~al.: The role of explainable ai in the context of the ai act. In: Proceedings of the 2023 ACM Conference on Fairness, Accountability, and Transparency. pp. 1139--1150 (2023)

\bibitem{poeta2023concept}
Poeta, E., Ciravegna, G., Pastor, E., Cerquitelli, T., Baralis, E.: Concept-based explainable artificial intelligence: A survey. arXiv preprint arXiv:2312.12936  (2023)

\bibitem{radford2021learning}
Radford, A., Kim, J.W., Hallacy, C., Ramesh, A., Goh, G., Agarwal, S., Sastry, G., Askell, A., Mishkin, P., Clark, J., et~al.: Learning transferable visual models from natural language supervision. In: International conference on machine learning. pp. 8748--8763. PMLR (2021)

\bibitem{ribeiro2016should}
Ribeiro, M.T., Singh, S., Guestrin, C.: " why should i trust you?" explaining the predictions of any classifier. In: Proceedings of the 22nd ACM SIGKDD international conference on knowledge discovery and data mining. pp. 1135--1144 (2016)

\bibitem{rudin2019stop}
Rudin, C.: Stop explaining black box machine learning models for high stakes decisions and use interpretable models instead. Nature machine intelligence  \textbf{1}(5),  206--215 (2019)

\bibitem{sawada2022concept}
Sawada, Y., Nakamura, K.: Concept bottleneck model with additional unsupervised concepts. IEEE Access  \textbf{10},  41758--41765 (2022)

\bibitem{selvaraju2017grad}
Selvaraju, R.R., Cogswell, M., Das, A., Vedantam, R., Parikh, D., Batra, D.: Grad-cam: Visual explanations from deep networks via gradient-based localization. In: Proceedings of the IEEE international conference on computer vision. pp. 618--626 (2017)

\bibitem{shwartz2017opening}
Shwartz-Ziv, R., Tishby, N.: Opening the black box of deep neural networks via information. arXiv preprint arXiv:1703.00810  (2017)

\bibitem{srivastava2024vlg}
Srivastava, D., Yan, G., Weng, L.: Vlg-cbm: Training concept bottleneck models with vision-language guidance. Advances in Neural Information Processing Systems  \textbf{37},  79057--79094 (2024)

\bibitem{info_plane}
Tishby, N., Pereira, F.C., Bialek, W.: The information bottleneck method. arXiv preprint physics/0004057  (2000)

\bibitem{skinlesions}
Tschandl, P., Rosendahl, C., Kittler, H.: The ham10000 dataset, a large collection of multi-source dermatoscopic images of common pigmented skin lesions. Scientific data  \textbf{5}(1), ~1--9 (2018)

\bibitem{veale2021demystifying}
Veale, M., Zuiderveen~Borgesius, F.: Demystifying the draft eu artificial intelligence act—analysing the good, the bad, and the unclear elements of the proposed approach. Computer Law Review International  \textbf{22}(4),  97--112 (2021)

\bibitem{cub}
Wah, C., Branson, S., Welinder, P., Perona, P., Belongie, S.: The caltech-ucsd birds-200-2011 dataset. Tech. Rep. CNS-TR-2011-001, California Institute of Technology (2011)

\bibitem{wang2023learning}
Wang, B., Li, L., Nakashima, Y., Nagahara, H.: Learning bottleneck concepts in image classification. In: Proceedings of the IEEE/CVF Conference on Computer Vision and Pattern Recognition. pp. 10962--10971 (2023)

\bibitem{yang2023language}
Yang, Y., Panagopoulou, A., Zhou, S., Jin, D., Callison-Burch, C., Yatskar, M.: Language in a bottle: Language model guided concept bottlenecks for interpretable image classification. In: Proceedings of the IEEE/CVF Conference on Computer Vision and Pattern Recognition. pp. 19187--19197 (2023)

\bibitem{tiny_imagenet}
Yao, L., Miller, J.: Tiny imagenet classification with convolutional neural networks. CS 231N  \textbf{2}(5), ~8 (2015)

\bibitem{yeh2020completeness}
Yeh, C.K., Kim, B., Arik, S., Li, C.L., Pfister, T., Ravikumar, P.: On completeness-aware concept-based explanations in deep neural networks. Advances in neural information processing systems  \textbf{33},  20554--20565 (2020)

\bibitem{yuksekgonul2022posthoc}
Yuksekgonul, M., Wang, M., Zou, J.: Post-hoc concept bottleneck models. In: ICLR 2022 Workshop on PAIR{\textasciicircum}2Struct: Privacy, Accountability, Interpretability, Robustness, Reasoning on Structured Data (2022), \url{https://openreview.net/forum?id=HAMeOIRD_g9}

\bibitem{zeiler2014visualizing}
Zeiler, M.D., Fergus, R.: Visualizing and understanding convolutional networks. In: Computer Vision--ECCV 2014: 13th European Conference, Zurich, Switzerland, September 6-12, 2014, Proceedings, Part I 13. pp. 818--833. Springer (2014)

\bibitem{zhang2018interpretable}
Zhang, Q., Wu, Y.N., Zhu, S.C.: Interpretable convolutional neural networks. In: Proceedings of the IEEE conference on computer vision and pattern recognition. pp. 8827--8836 (2018)

\end{thebibliography}


\begin{thebibliography}{8}
\bibitem{ref_article1}
Author, F.: Article title. Journal \textbf{2}(5), 99--110 (2016)

\bibitem{ref_lncs1}
Author, F., Author, S.: Title of a proceedings paper. In: Editor,
F., Editor, S. (eds.) CONFERENCE 2016, LNCS, vol. 9999, pp. 1--13.
Springer, Heidelberg (2016). \doi{10.10007/1234567890}

\bibitem{ref_book1}
Author, F., Author, S., Author, T.: Book title. 2nd edn. Publisher,
Location (1999)

\bibitem{ref_proc1}
Author, A.-B.: Contribution title. In: 9th International Proceedings
on Proceedings, pp. 1--2. Publisher, Location (2010)

\bibitem{ref_url1}
LNCS Homepage, \url{http://www.springer.com/lncs}, last accessed 2023/10/25
\end{thebibliography}
%

\clearpage
\appendix

\section{Derivation details}
\label{app:derivation}
Since $c$ is unknown, we estimate it leveraging the observable variables $x$ and $y$. Formally, this requires modeling the posterior distribution $p(c\mid x,y) = \frac{p(c,x,y)}{p(x,y)}$. However, direct computation of this posterior is infeasible due to the intractability of the denominator, where $p(x,y) = \int p(x,c,y) \, dc$. To address this, we approximate the true posterior $p(c\mid x,y)$ with a variational distribution $q(\rev{c}\mid x)$, conditioned only on $x$ because $y$ is unobservable at test time. In order to find a good posterior approximation we minimize KL divergence among the approximated $q(c\mid x)$ and true $p(c\mid x,y)$ posterior. For simplicity in the notation, we will refer to this term using $KL(q||p) = KL(q(c|x)||p(c|x,y)) $:
{\scriptsize
\begin{align}
KL(q||p) &= \int q(c|x)\text{log }\frac{\rev{q}(c|x)}{p(c|x,y)} dc \\
    &= \int q(c|x)\text{log }q(c|x)dc - \int q(c|x)\text{log }p(c|x,y)dc \\
    &= E_q \left[\text{log }q(c|x)\right] - E_q\left[\text{log }\frac{p(x,y,c)}{p(x,y)}\right]\\
    &= E_q \left[\text{log }q(c|x)\right] - E_q\left[\text{log }p(x,y,c)\right] + E_q\left[\text{log }p(x,y)\right]\\
    &= E_q \left[\text{log }q(c|x)\right] - E_q\left[\text{log }p(x,y,c)\right] + \text{log }p(x,y)
\end{align}
}
Rearranging the terms yields:

{\scriptsize
\begin{equation*}
\begin{aligned}
    \underbrace{E_q\left[\text{log }p(x,y,c)-\text{log }q(c|x)\right]}_{\text{ELBO}} = \text{log }p(x,y) - KL(q||p) 
\end{aligned}
\end{equation*}
}
The expression on the left-hand side of the equation is referred to as the Evidence Lower Bound (ELBO), the maximization of which allows to approximate the joint distribution $p(x,y)$, while reducing the KL divergence between the approximate posterior and a given prior $p(c)$.

{\scriptsize
\begin{align}
    \text{ELBO} &= E_q\left[\text{log }p(x,y,c)-\text{log }q(c|x)\right]\\
    &= E_q\left[\text{log }p(x,y|c)\right] + E_q\left[\text{log }\frac{p(c)}{q(c|x)}\right]\\
    &= E_q\left[\text{log }p(x,y|c)\right] - KL(q(c|x)||p(c))
\end{align}
}

We now assume conditional independence of $x$ and $y$ with respect to $c$. This implies that the first term on the right-hand side of the equation can be rewritten as $p(x,y|c) = p(x|c)p(y|c)$, leading to the following reformulation of the ELBO:

{\scriptsize
\begin{equation}
    \text{ELBO} = \overbrace{E_q[\text{log }p(x|c)]}^{\text{Representativity}} + \overbrace{E_q[\text{log }p(y|c)]}^{\text{Completeness}} - \overbrace{KL(q(c|x)||p(c))}^{\text{Aignement}}
\end{equation}
}


\section{ViT-base-patch32 Results}
\label{app:app_vit}
In this appendix, we report the results in terms of generalization performance across the evaluated datasets for the different methodologies using ViT-base-patch32 as the backbone. Results for ProtoPNet~\cite{chen2019looks} are not included, as it was not designed to work with a transformer-based backbone. Compared to the results obtained with ResNet-18, the performance are generally higher, reflecting the increased complexity of the backbone. Nonetheless, even in this case, \XXX{} consistently provides the highest generalization accuracy, closing the gap with end-to-end black-box models. The only exception is MNIST Even/Odd, where BotCL achieves an accuracy approximately 0.3\% higher. On more complex datasets such as Tiny ImageNet, \XXX{} can outperform the best baseline model by more than 5\%.

\begin{table*}[h]
    \centering
    \caption{Task accuracy of the different methodologies using Vit-base-patch32 as backbone.}
    \label{tab:vit_task_acc}
    \resizebox{\textwidth}{!}{%
    \begin{tabular}{@{}lccccccc@{}}
        \toprule
        & \textbf{MNIST E/O} & \textbf{MNIST Addition} & \textbf{CIFAR10} & \textbf{CIFAR100} & \textbf{Tiny ImageNet} & \textbf{Skin lesions} & \textbf{CUB200} \\ \midrule
        \textbf{Method \textbackslash Conc.} & 10 & 10 & 15 & 20 & 30 & 14 & 112\\ 
        \midrule
        E2E & 98.42 ± 0.02 & 64.02 ± 0.03 & 94.62 ± 0.05 & 80.34 ± 0.36 & 73.01 ± 0.06 & 89.83 ± 1.12 & 78.65 ± 0.73\\ 
        \midrule
        BotCL (recon) &  78.43 ± 2.43 &  13.15 ± 4.12 & 30.15 ± 3.42 & 8.12 ± 0.12 &  5.41 ± 2.11 & 82.02 ± 1.45 & 35.78 ± 3.89\\
        BotCL (contr) & \textbf{98.12 ± 0.06}  &  47.52 ± 2.42 &  91.51 ± 0.23 & 68.90 ± 1.24 &  64.01 ± 2.02 & 86.11 ± 1.11 & 75.46 ± 0.42\\
        SENN &  96.70 ± 0.13 &  48.10 ± 0.86 &  91.97 ± 0.10 &  74.23 ± 0.62 & 66.32 ± 0.81 & 85.31 ± 0.91 & 70.01 ± 0.54\\
        LF-CBM &  97.47 ± 0.06  &  53.44 ± 0.60  &  92.06 ± 1.11  &  62.89 ± 0.92 & 65.06 ± 1.06 &   52.13 ± 0.55  &  77.05 ± 0.32 \\
        \textbf{\XXX{} (ours)} & 97.80 ± 0.19 & \textbf{61.25 ± 0.53} & \textbf{94.06 ± 0.11} & \textbf{78.66 ± 0.30} & \textbf{71.93 ± 0.23} & \textbf{89.43 ± 0.04} & \textbf{78.05 ± 0.31}\\ \bottomrule
    \end{tabular}
    }
\end{table*}

\section{Impact of Concept Embedding Size}  
\label{app:abl_emb}  

The dimensionality \( d \) of the concept embedding \( \mathbf{c}_j \) determines the amount of information that can be transmitted through each concept. Consequently, increasing the embedding size directly influences both the reconstruction quality (\textit{Representativity}) and the task performance (\textit{Completeness}). Specifically, a larger \( d \) allows the decoder and classifier to leverage more information for reconstructing the input sample \( x \) and predicting the correct label \( y \).  

In this appendix, we analyze how task performance varies with different concept embedding sizes.  

\begin{figure*}[h]
    \centering
    \includegraphics[width=0.5\textwidth]{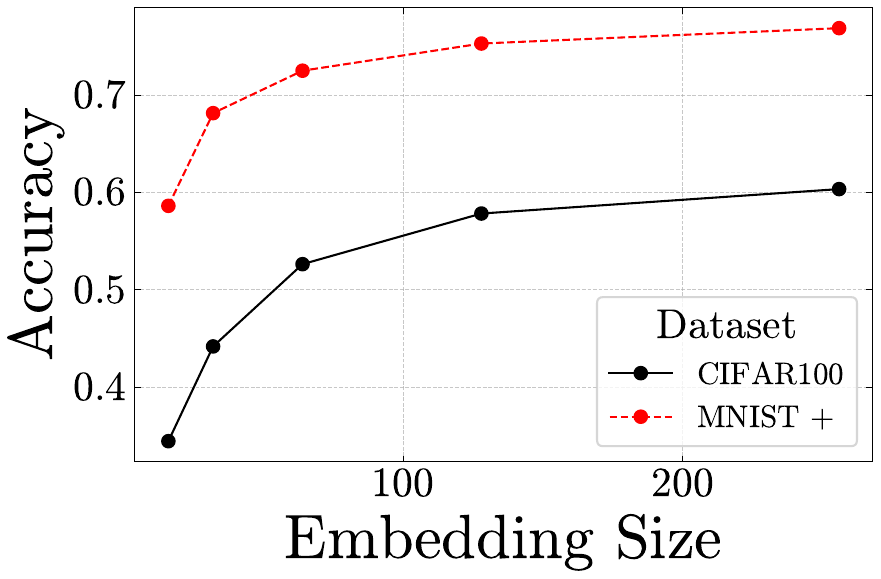}
    \caption{Effect of concept embedding size on task performance. As the embedding size increases, accuracy improves on both datasets until it reaches a plateau.}
    \label{fig:ab_emb}
\end{figure*}

As shown in Fig.~\ref{fig:ab_emb}, increasing the embedding size leads to improved accuracy on both datasets, eventually reaching a plateau. To balance high accuracy with a manageable number of parameters, we set \( d = 128 \) for our experiments.

\section{Dataset details}
\label{app:app_dataset}
In this work, we utilize seven datasets: MNIST~\cite{mnist}, MNIST Addition, CIFAR-10~\cite{cifar10_100}, \hbox{CIFAR-100}~\cite{cifar10_100}, Tiny ImageNet~\cite{tiny_imagenet}, Skinlesions~\cite{skinlesions} and \hbox{CUB-200}~\cite{cub}. 
Images had to be resized to $224\times224$ and normalized to match the input specifics of the backbones. For CIFAR-10, CIFAR-100, Tiny ImageNet, Skinlesions and CUB-200 data augmentation was applied. More precisely, we applied: random horizontal flip with $p=0.5$; random resized crop with scale=(0.08, 1.0), starting from a rescaled image size of $280\times280$; random rotation with degree=0.1. For each dataset we performed a training/validation split by randomly selecting 10\% of the samples in the training-set to create the validation-set. Consequently, for each experiment, we select the model which performed better on the validation-set. MNIST Even/Odd contains \num{60000} training images and \num{10000} test images of handwritten digits, each of size $28\times28$ and in grayscale, that must be classified as even or odd.  The images were converted to three channels to align with the input requirements of the models. MNIST Addition is created by pairing two MNIST digits and assigning a label equal to the sum of their individual labels. It contains \num{60000} training samples and \num{10000} test samples, matching the original MNIST dataset. Each input consists of a grayscale image of size $56\times28$, which is converted to a three-channel image to match the model’s input requirements. CIFAR-10 comprises \num{60000}, 
$32\times32$ color images across 10 classes, with \num{50000} images for training and \num{10000} for testing. CIFAR-100 is identical to CIFAR-10 in terms of image size and number of channels but it includes 100 classes (600 images per class, 500 for training and 100 for testing). Tiny Imagenet consists of \num{100000} $64\times64$ color images across 200 classes, with 500 training images and 50 testing images per class. Skin Lesions is a dataset of \num{10015} dermatoscopic images spanning 14 skin lesion categories, which we use as concepts. For classification, we group these into four macro classes: \textit{Healthy}, \textit{Benign}, \textit{Malignant}, and \textit{Infectious Diseases}. CUB-200 is a fine-grained bird classification dataset containing \num{11788} images of 200 species, split into \num{5994} for training and \num{5794} for testing. Each class has varying numbers of images, and for our experiments, we selected the 112 most balanced concepts.

\section{User Study Details}
\label{app:survey}
In this appendix, we provide some additional details regarding the user study we conducted. The goal was to compare the \textit{Plausibility} of the concepts extracted by the model by asking users to select an image that may belong to a concept and identify the intruder image in a set of images representing a single concept. In the same study, we also evaluated the \textit{Human Understanding} of the concepts by asking participants to name a set of images representing each concept. This user study involved 72 users, each answering 18 questions divided into three tasks. Half of these questions were related to BotCL~\cite{wang2023learning}, and the other half to our method. 

\begin{figure}[h]
    \centering
        \includegraphics[width=0.7\linewidth]{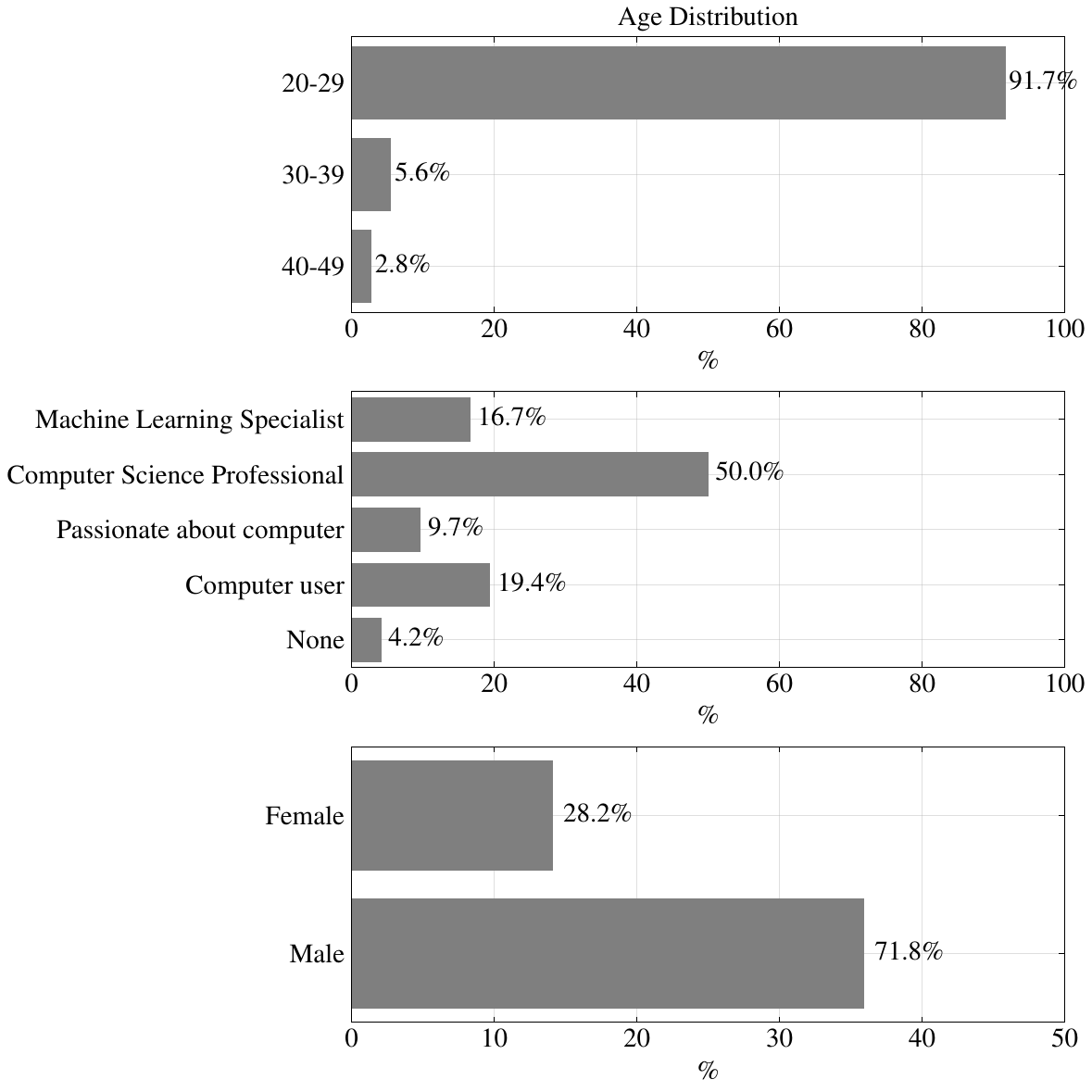} 
        \caption{Characterization of the users involved in the user study.}
    \label{fig:survey_info}
\end{figure}

In Fig.~\ref{fig:survey_info}, we present some statistics from the survey, including the gender distribution, which shows a predominance of male-identifying users, the user age distribution, which is polarized in the 20-29 range, and their level of expertise, which is spread across 5 levels. All participants in the user study were informed in advance about the content and purpose of the study.

\begin{figure*}[h]
    \centering
    \begin{subfigure}[b]{0.45\textwidth}
        \includegraphics[width=\textwidth]{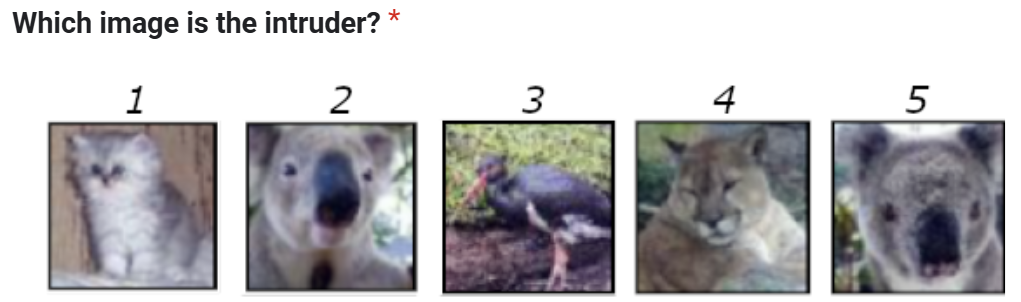}
        \caption{Example of a type 1 question. The intruder, image 3, can be identified either as the only image depicting a bird or as the only animal with a black coat.}
        \label{fig:survey_intruder}
    \end{subfigure}\hfill
    \begin{subfigure}[b]{0.45\textwidth}
        \includegraphics[width=\textwidth]{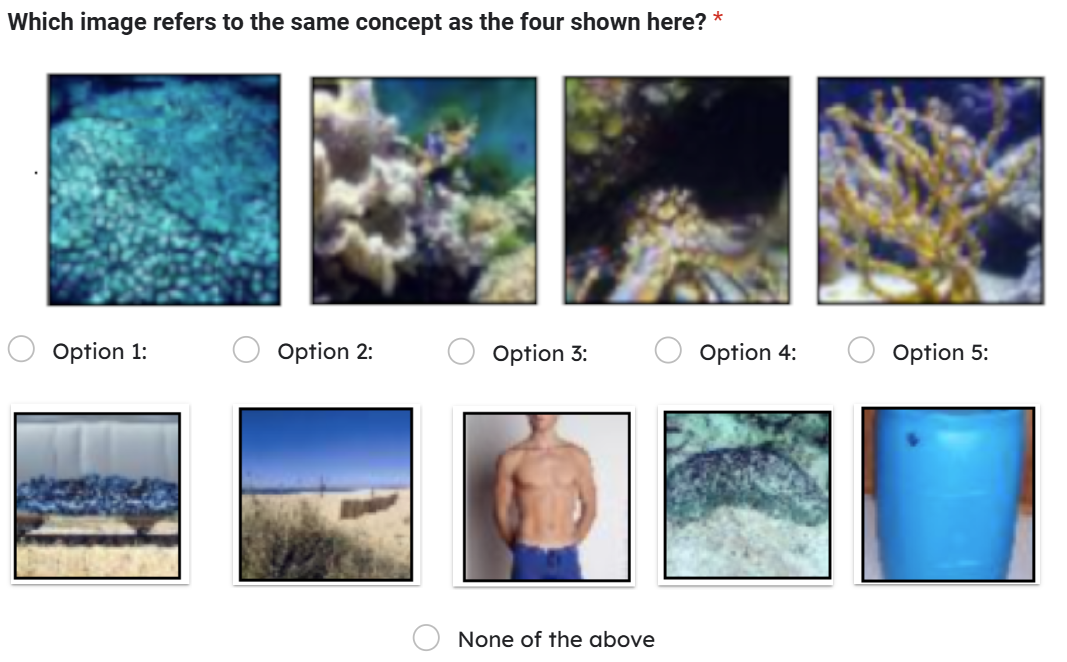}
        \caption{Example of a type 2 question. Since all the images above depict corals, the correct answer is the fourth image.}        
        \label{fig:survey_concept}
    \end{subfigure}    
    \begin{subfigure}[b]{0.45\textwidth}
        \vspace{0.5cm}
        \includegraphics[width=\linewidth]{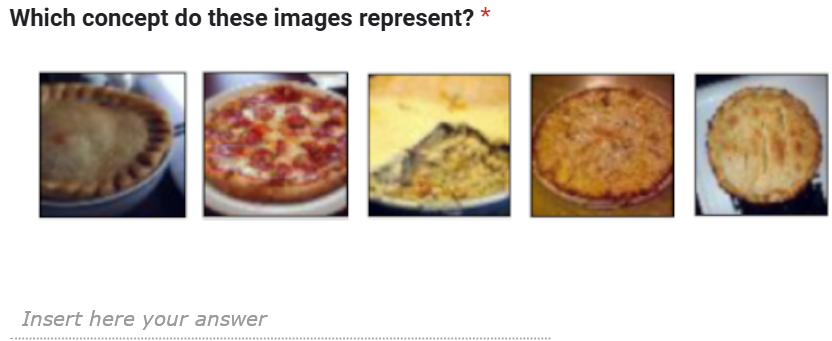} 
        \caption{Example of a type 3 question. All the images depict various foods with a round shape, so users are expected to provide representative keywords such as "food," "pizza," "pie," or "circular-shaped food."}
        \label{fig:survey_tag}
    \end{subfigure}
    \caption{The figure presents an example for each type of question displayed to users in the study. All examples in the figure were sourced from the dictionaries generated by \XXX.}
    \label{fig:two_top_one_bottom}
\end{figure*}

An example of the first type of question presented to the users is depicted in Fig.~\ref{fig:survey_intruder}. The user is asked to find the intruder image among the 5 images presented. The images are extracted from the dictionaries of the various datasets, all of which are included in this paper. In all the questions presented, there is always one and only one intruder. If the user is unable to choose, they can select the option `None`, which is counted as an incorrect answer. In Fig.~\ref{fig:survey_concept}, the second type of question is shown. In this case, given four images belonging to a concept present in one of the dictionaries, the user is asked to select, among the options provided, another image related to the same concept. In all the questions, there is one correct answer, while all incorrect options are taken from the same dictionary but correspond to different concepts. If the user is unable to choose, they can select the option `None of the above`, which is counted as an incorrect answer. In the last type of question, shown in Fig.~\ref{fig:survey_tag}, the user is asked to input, using one or two words, the concept being represented. If they are unable to name the concept they are instructed to write "no pattern".

\section{Detailed Task Accuracy}
\label{app:task_acc}
In this appendix, we provide in Table~\ref{tab:task_acc} the detailed generalization performance of \XXX, the baselines, and the end-to-end black-box models reported in Figure~\ref{fig:task_acc} and discussed in Section~\ref{subsec:generalization}.

\begin{table*}[h]
    \centering
    \caption{Task accuracy of the different methodologies using ResNet-18 as backbone.}
    \resizebox{\textwidth}{!}{%
    \begin{tabular}{@{}lccccccc@{}}
        \toprule
        & \textbf{MNIST Even/Odd} & \textbf{MNIST Addition} & \textbf{CIFAR10} & \textbf{CIFAR100} & \textbf{Tiny ImageNet} & \textbf{Skin Lesions} & \textbf{CUB200} \\ \midrule
        E2E & 98.58 ± 0.17 & 73.51 ± 0.42 & 81.50 ± 0.16 & 59.51 ± 0.11 & 57.61 ± 0.44 & 82.17 ± 0.22 & 67.96 ± 0.34 \\ \midrule
        BotCL (recon) & 88.06 ± 7.45 & 17.65 ± 10.98 & 67.32 ± 8.48 & 7.53 ± 0.06 & 6.64 ± 0.09 & 72.63 ± 2.48 & 20.11 ± 2.38 \\
        BotCL (contr) & \textbf{98.01 ± 0.29} & 44.83 ± 0.02 & 77.11 ± 0.51 & 33.72 ± 0.59 & 33.59 ± 0.71 & 81.71 ± 0.76 & 51.12 ± 0.94 \\
        SENN & 96.13 ± 0.03 & 59.62 ± 0.01 & 75.60 ± 0.09 & 43.46 ± 0.01 & 38.17 ± 0.02 & 76.77 ± 1.26 & 44.84 ± 1.26 \\
        ProtoPNet & 97.47 ± 0.23 & 70.48 ± 0.04 & 75.26 ± 3.81 & 50.81 ± 1.95 & 39.59 ± 0.97 & 81.78 ± 0.23 & 55.80 ± 0.86 \\
        LF-CBM & 97.68 ± 0.32 & 56.65 ± 0.49 & 78.68 ± 0.16 & 38.76 ± 0.46 & 43.07 ± 0.20 & 45.99 ± 0.01 & 58.79 ± 1.03 \\ \midrule
        \textbf{\XXX{} (ours)} & 97.25 ± 0.97 & \textbf{75.44 ± 0.04} & \textbf{81.33 ± 0.15} & \textbf{57.92 ± 0.56} & \textbf{57.79 ± 0.17} & \textbf{81.82 ± 0.61} & \textbf{60.52 ± 1.01} \\ \bottomrule
    \end{tabular}%
    }
    \label{tab:task_acc}
\end{table*}

\section{CAS Concept Alignment}
\label{app:cas_concept_align}

In this appendix, we provide in Table~\ref{tab:concepts_cas} the detailed Concept Alignment Score~\cite{zarlenga2022concept} for candidate concepts with respect to existing human representations. As already presented in Section~\ref{subsec:interpret} for the macro F1-score, CAS similarly highlights how \XXX{} concepts are more aligned with human-defined representations. Specifically, for all datasets, \XXX{} outperforms the baselines by at least 0.40, with an improvement exceeding 0.60 on MNIST Addition.

\begin{table*}[h]
    \centering
    \caption{CAS for candidate concepts with respect to existing human-representations}
    \resizebox{\textwidth}{!}{%
    \begin{tabular}{@{}lccccc@{}}
        \toprule
        & \textbf{MNIST Even/Odd} & \textbf{MNIST Addition} & \textbf{CIFAR100} & \textbf{Skin lesion} & \textbf{CUB200}\\ \midrule
        BotCL (recon)                    & 0.07 \footnotesize{±$\le$0.01} & 0.05 \footnotesize{±$\le$0.01} & 0.09 \footnotesize{± 0.03} & 0.07 \footnotesize{±$\le$0.01} & 0.07 \footnotesize{± 0.02} \\
        BotCL (contr)                    & 0.13 \footnotesize{± 0.01} & 0.12 \footnotesize{± 0.01} & 0.17 \footnotesize{± 0.02} & 0.11 \footnotesize{± 0.01} & 0.08 \footnotesize{± 0.02}\\
        SENN                             & 0.14 \footnotesize{± 0.01} & 0.27 \footnotesize{± 0.02} & 0.08 \footnotesize{± 0.01} & 0.22 \footnotesize{± 0.02} & 0.13 \footnotesize{± 0.02}\\
        ProtoPNet                        & 0.05 \footnotesize{±$\le$0.01} & 0.08 \footnotesize{± 0.01} & 0.09 \footnotesize{± 0.01} & 0.12 \footnotesize{± 0.02} & 0.15 \footnotesize{± 0.01}\\
        LF-CBM                           & 0.03 \footnotesize{±$\le$0.01} & 0.01 \footnotesize{±$\le$0.01} & 0.03 \footnotesize{±$\le$0.01} & 0.01 \footnotesize{±$\le$0.01} & 0.01 \footnotesize{±$\le$0.01}\\
        \midrule
        \textbf{\XXX{} (ours)}           & \textbf{0.86 \footnotesize{±$\le$0.01}} & \textbf{0.81 \footnotesize{±$\le$0.01}} & \textbf{0.67 \footnotesize{±$\le$0.01}} & \textbf{0.65 \footnotesize{±$\le$0.01}} & \textbf{0.63 \footnotesize{±$\le$0.01}}\\ \bottomrule
    \end{tabular}
    }
    \label{tab:concepts_cas}
\end{table*}

\section{Additional Qualitative Results}
In this appendix, we present additional qualitative results, showcasing both the dictionaries constructed by our method for each dataset and some sample explanations. 

\subsection{Discovered Dictionaries}
\label{app:dictionary}
\begin{figure*}[h]
    \centering
        \includegraphics[width=\linewidth]{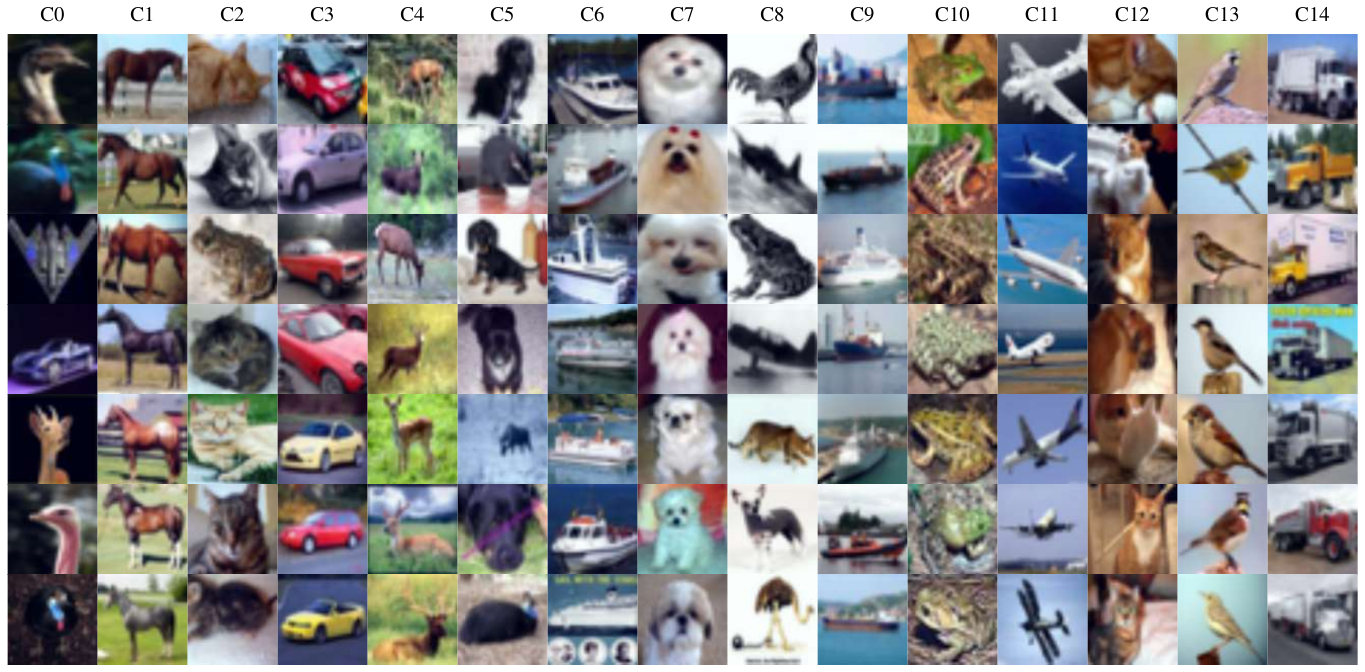} 
        \caption{CIFAR10 dictionary of concepts.} 
    \label{fig:cifar10_dict}
\end{figure*}

To illustrate the dictionaries, for each concept, we show the 7 images of the training set that mostly activated the given concept (whose concept embedding was more aligned with the concept prototype). The first dictionary we present is the one corresponding to CIFAR-10, shown in Fig~\ref{fig:cifar10_dict}. The 15 concepts are well-distinguished, and among them, we mention $C1$, which contains images of horses, or $C14$, which represents `trucks'. More specific are concept $C0$, which depicts images of animals or objects with similar dark backgrounds, and concept $C8$, which instead depicts animals on light backgrounds.

\begin{figure*}[h]
    \centering
        \includegraphics[width=\linewidth]{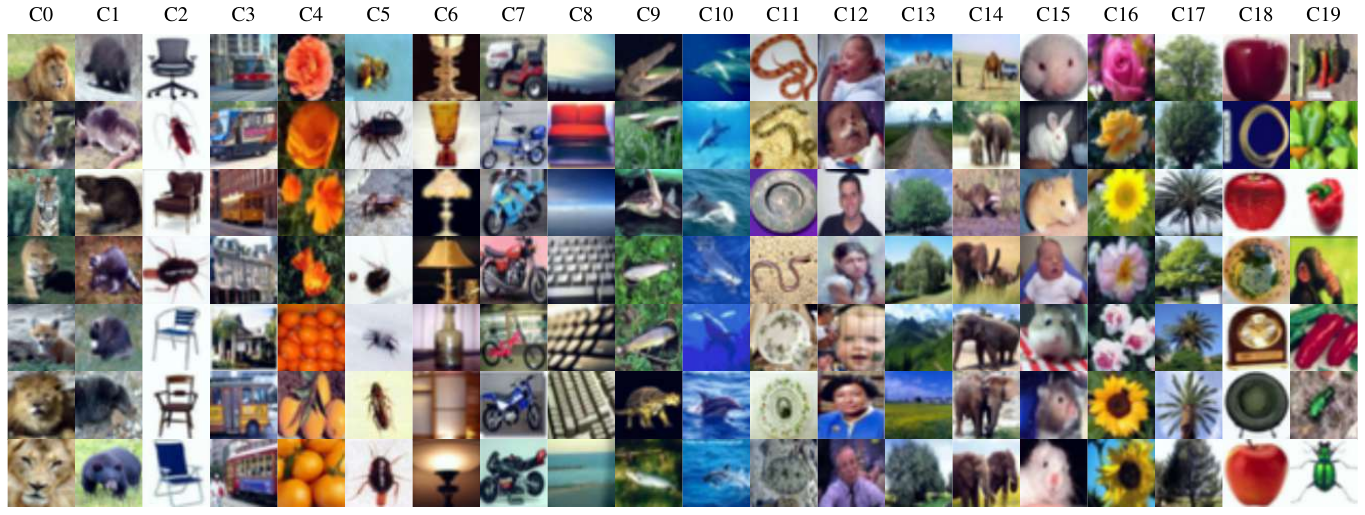} 
        \caption{CIFAR-100 dictionary of concepts.} 
    \label{fig:cifar100_dict}
\end{figure*}

The second dictionary we analyze is the one corresponding to CIFAR-100, shown in Fig~\ref{fig:cifar100_dict}. Some of the 20 represented concepts appear more specific, such as concept $C12$, which groups images of people, while others are more general, such as concept $C13$, which represents `landscapes with trees'. Among these types of concepts is concept $C2$, which group objects with a white background. Many of them also seem to align well with CIFAR-100 superclasses, such as concept $C0$, which represents `large carnivores`, or concept $C10$, which represents `aquatic animals`, or even concept $C16$, which represents `flowers`.

\begin{figure*}[h]
    \centering
        \includegraphics[width=\linewidth]{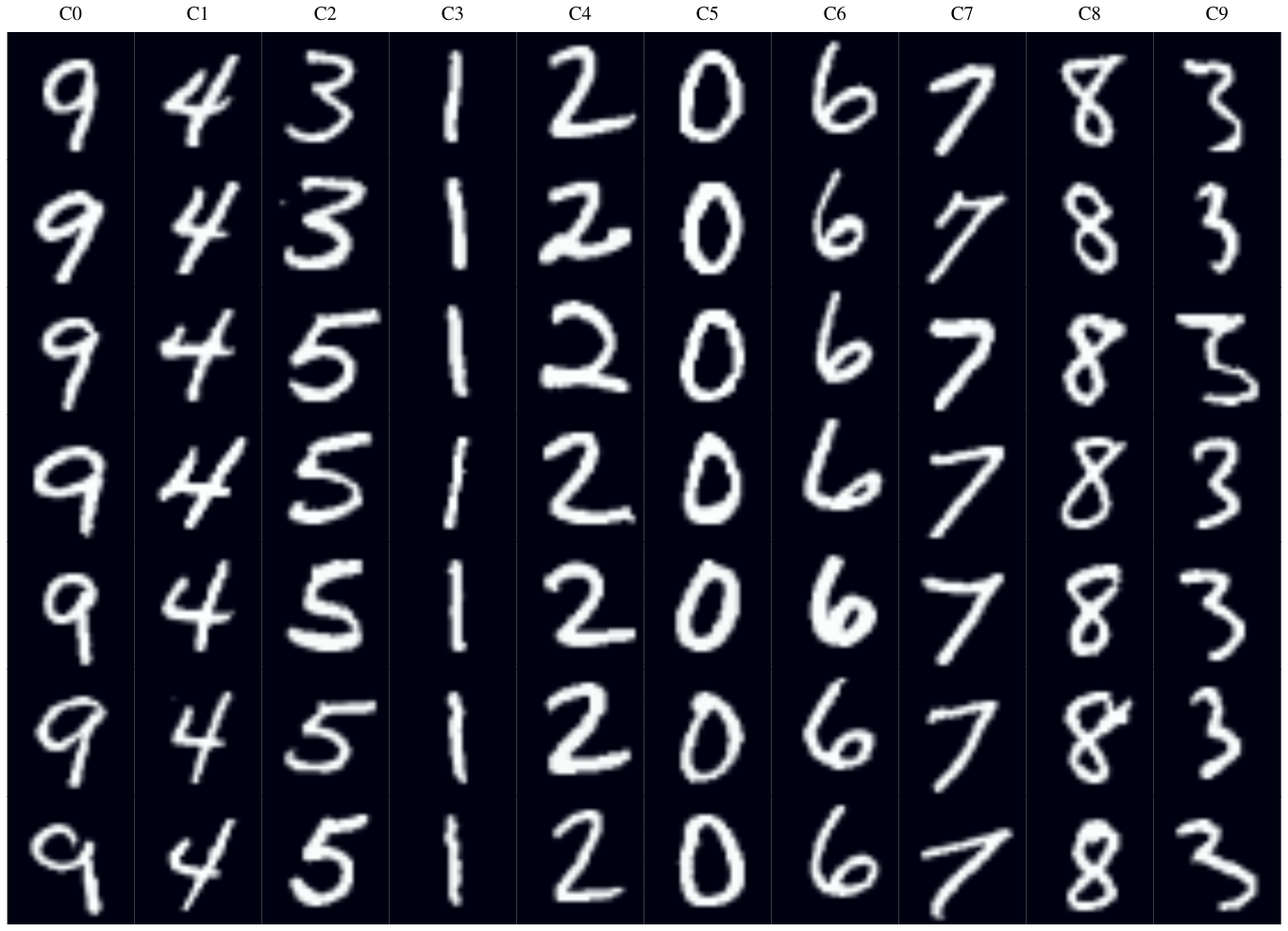} 
        \caption{MNIST Even/Odd dictionary of concepts.} 
    \label{fig:mnist_eo_dict}
\end{figure*}

The MNIST dictionary shown in Fig~\ref{fig:mnist_eo_dict} is straightforward to analyze as the concepts align perfectly with the ten concepts humans would use to determine if a number is even or odd: the digits themselves. This explains the high level of alignment reflected by Tab.~\ref{tab:concepts_f1} and Tab.~\ref{tab:concepts_cas}. 

\begin{figure*}[h]
    \centering
        \includegraphics[width=\linewidth]{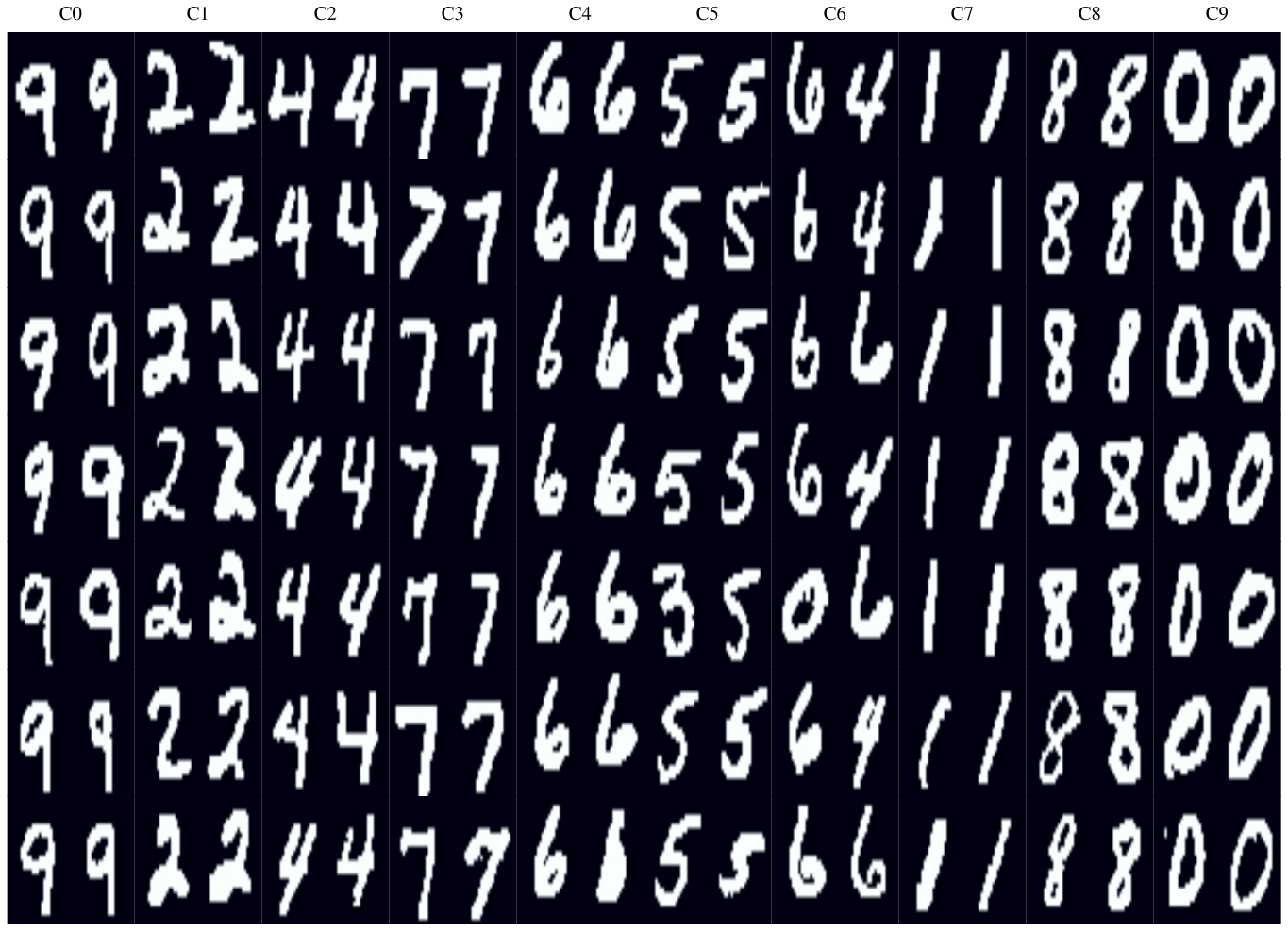} 
        \caption{MNIST Addition dictionary of concepts.} 
    \label{fig:mnist_sum_dict}
\end{figure*}

A similar phenomenon occurs with the concept dictionary of MNIST Addition, shown in Fig~\ref{fig:mnist_sum_dict}. In this case, even though the task is different (i.e., predicting the sum of two numbers), the concept discovered by \XXX\ aligns perfectly with the digits. This result is further supported quantitatively by Tab.~\ref{tab:concepts_f1} and Tab.~\ref{tab:concepts_cas}. Taking $C0$ as an example, which corresponds to the digit 9, it is noteworthy that the images most strongly activating this concept consistently feature the digit 9 appearing twice, which holds true across all the dictionary.

\subsection{Explanations}
\label{app:explanations}
The explanations are organized as follows: the original image appears in the top-left corner. A vertical bar plot, located at the top-right, displays the importance of each concept for classifying a specific category (indicated in the far top-right corner of the figure). The x-axis of the bar plot shows the original image with the corresponding Grad-CAM applied to a concept. Below the bar plot, four images from the dictionary that correspond to the concept are displayed. 

\begin{figure*}[h]
    \centering
    \begin{minipage}{0.45\textwidth}
        \centering
        \includegraphics[width=\linewidth]{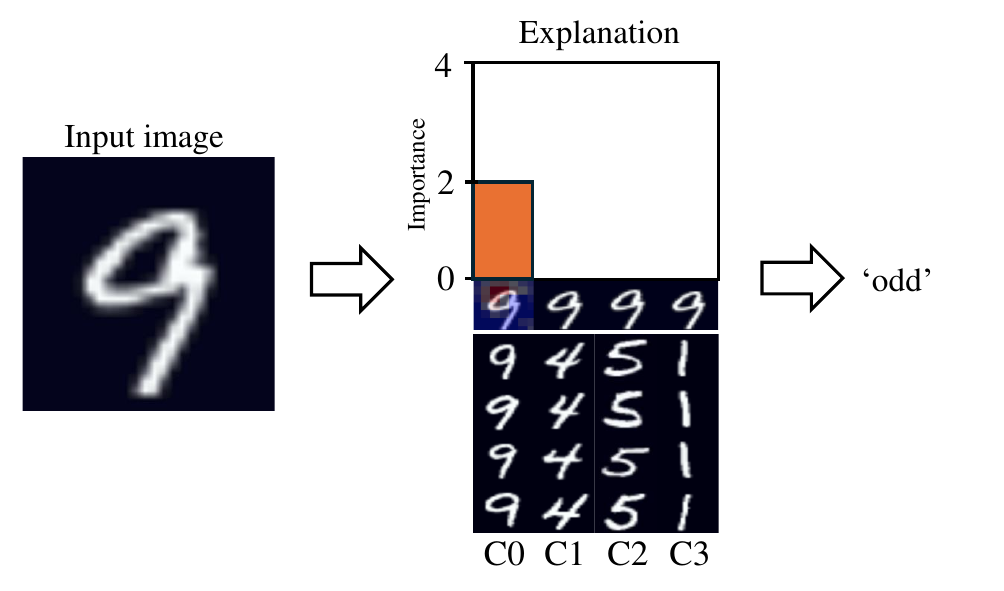}
        \caption{Example belonging to class "odd" extracted form the MNIST-Evan/Odd dataset.}
        \label{fig:MNIST_odd}
    \end{minipage}\hfill
    \begin{minipage}{0.45\textwidth}
        \centering
        \includegraphics[width=\linewidth]{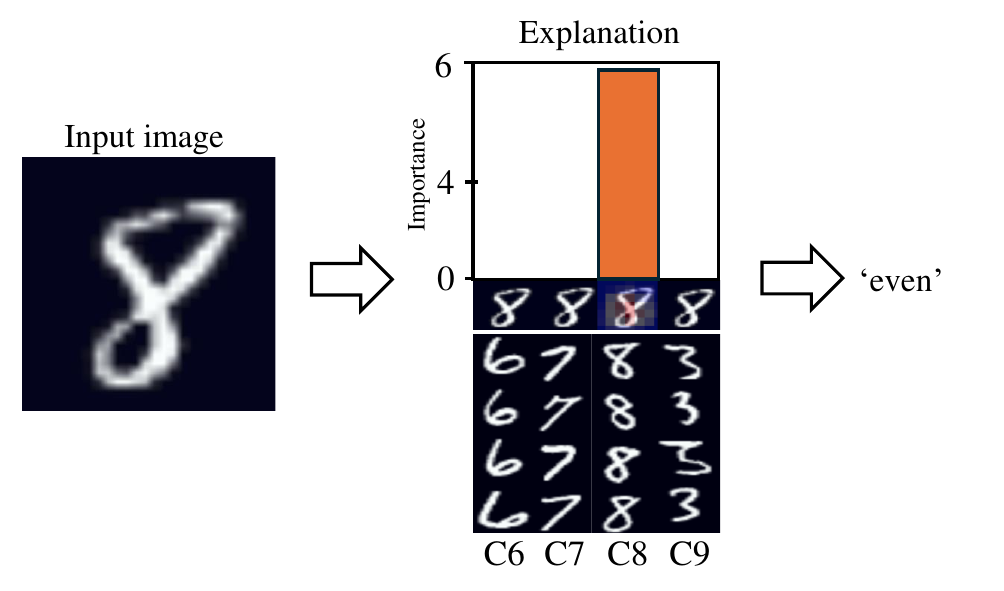}
        \caption{Example belonging to class "even" extracted form the MNIST-Evan/Odd dataset.}
        \label{fig:MNIST_even}
    \end{minipage}
    \label{fig:mnist_eo_explain}
\end{figure*}

The explanations related to MNIST Even/Odd, shown in Fig~\ref{fig:MNIST_odd} and Fig~\ref{fig:MNIST_even}, are clear and confirm the hypothesis that the model uses the~10 digits to correctly predict whether the input image represents an even or odd number.

\begin{figure*}[h]
    \centering
    \begin{minipage}{0.45\textwidth}
        \centering
        \includegraphics[width=\linewidth]{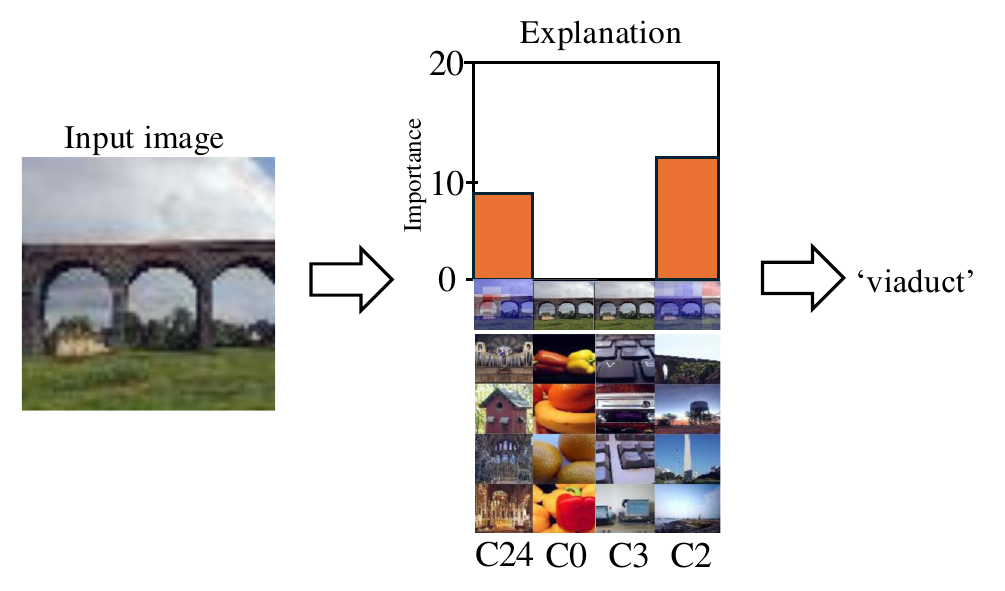}
        \caption{Example belonging to class "viaduct" extracted form the Tiny Imagenet dataset.}
        \label{fig:viaduct}
    \end{minipage}\hfill
    \begin{minipage}{0.45\textwidth}
        \centering
        \includegraphics[width=\linewidth]{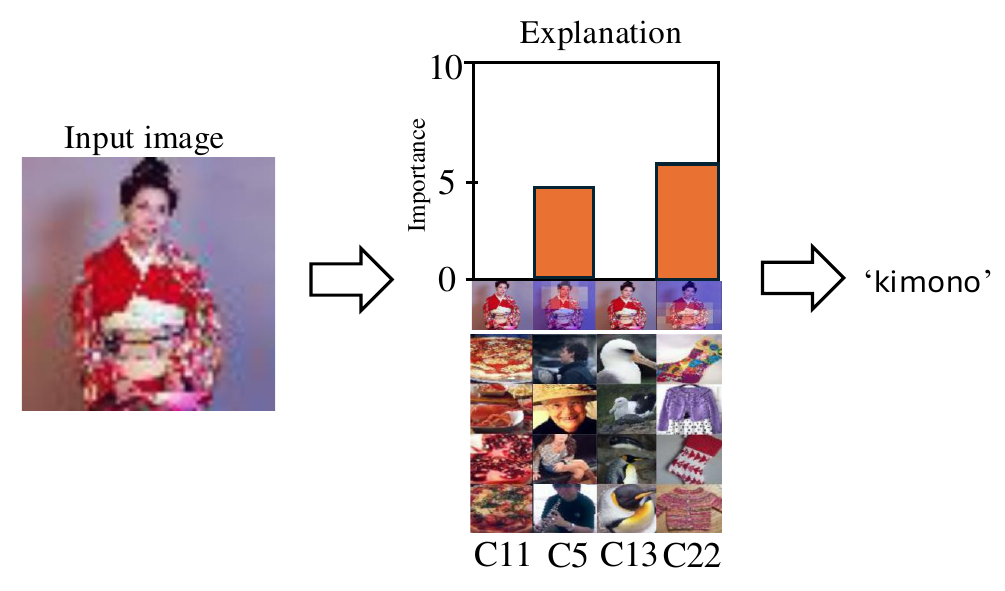}
        \caption{Example belonging to class "kimono" extracted form the Tiny Imagenet dataset.}
        \label{fig:asian_woman}
    \end{minipage}
\end{figure*}

A more complex example is presented in Fig~\ref{fig:viaduct}. Here, the input image depicts a viaduct, and the model correctly classifies the image by leveraging concept $C6$ (representing architectures/constructions) and concept $C10$ (representing the sky). Grad-CAMs reveal that $C24$ focuses on a specific section of the bridge, while $C2$ highlights the sky. On the other hand, Fig.~\ref{fig:asian_woman} illustrates an Asian woman wearing a kimono. The activated concepts in this case are $C5$, associated with human faces, and $C22$, related to clothing. The model, by recognizing the presence of an Asian woman and the clothing she is wearing, correctly predicts the label "kimono".

\begin{figure*}[h]
    \centering
    \begin{minipage}{0.45\textwidth}
        \centering
        \includegraphics[width=\linewidth]{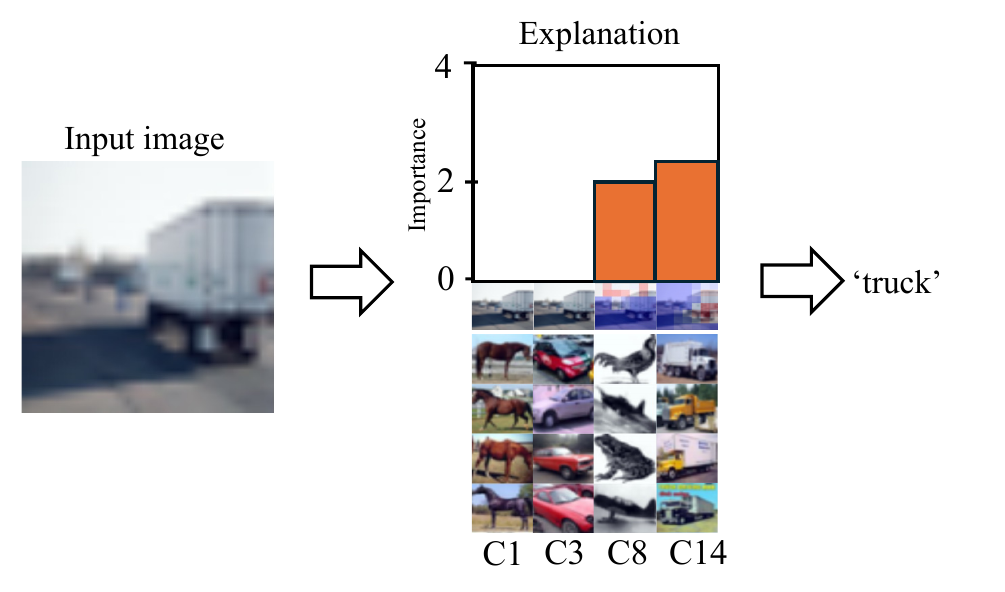}
        \caption{Example belonging to class "truck" extracted form the CIFAR-10 dataset.}
        \label{fig:truck}
    \end{minipage}\hfill
    \begin{minipage}{0.45\textwidth}
        \centering
        \includegraphics[width=\linewidth]{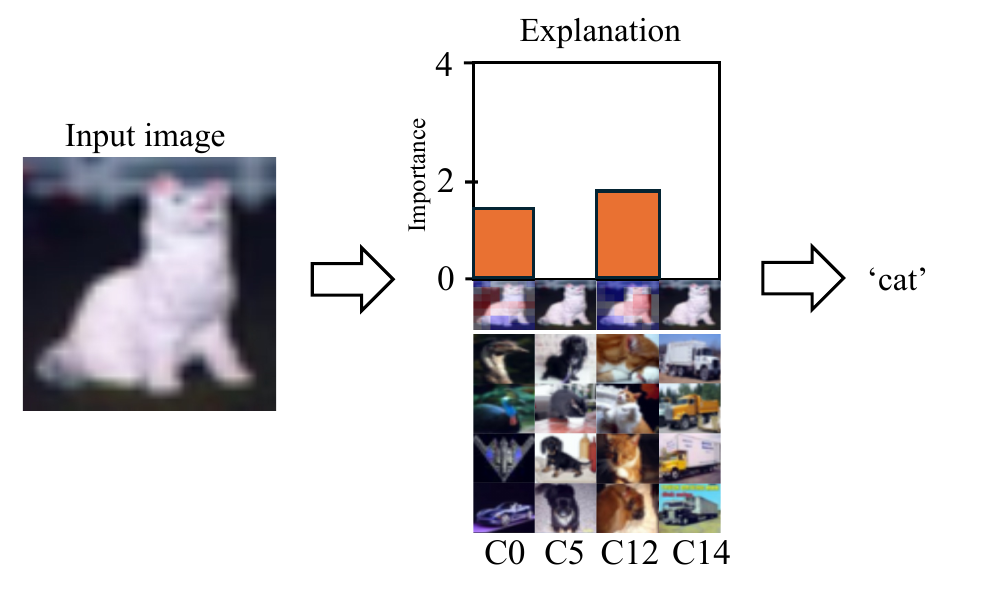}
        \caption{Example belonging to class "cat" extracted form the CIFAR-10 dataset.}
        \label{fig:cat}
    \end{minipage}
\end{figure*}

The examples in Fig.\ref{fig:truck} and Fig.\ref{fig:cat} illustrate how the \XXX\ in this case utilizes concepts that focus primarily on the object or animal, alongside another concept that highlights the background. In Fig.\ref{fig:truck}, concept $C8$ focuses on the white background, represented by the sky for the specific example, while concept $C14$ emphasizes the tires. Similarly, in Fig.\ref{fig:cat}, concept $C0$ focuses on the black background, and concept $C12$ targets the facial features of the cat.

\begin{figure*}[h]
    \centering
    \begin{minipage}{0.45\textwidth}
        \centering
        \includegraphics[width=\linewidth]{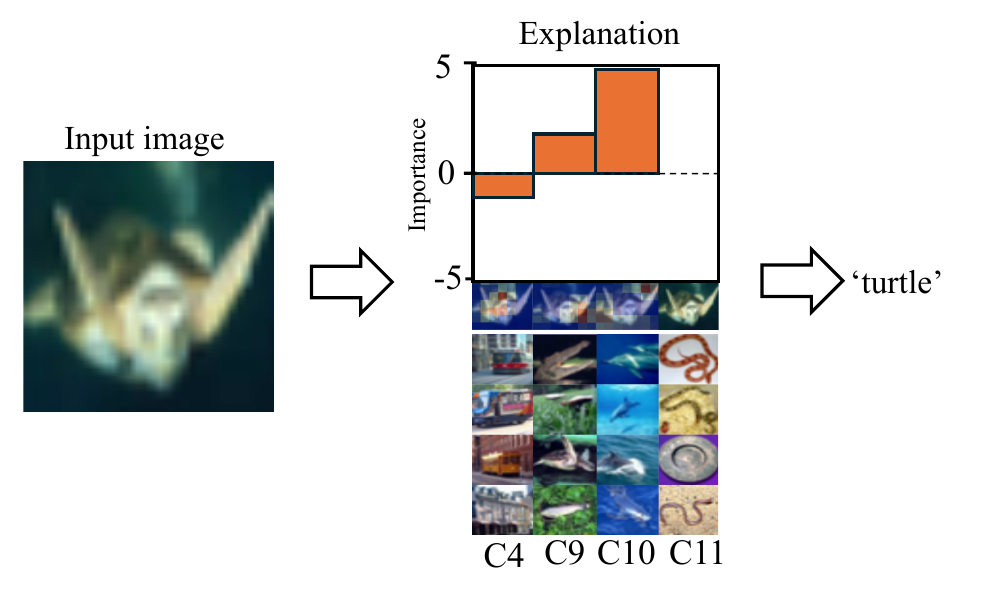}
        \caption{Example belonging to class "turtle" extracted form the CIFAR-100 dataset.}
        \label{fig:turtle}
    \end{minipage}\hfill
    \begin{minipage}{0.45\textwidth}
        \centering
        \includegraphics[width=\linewidth]{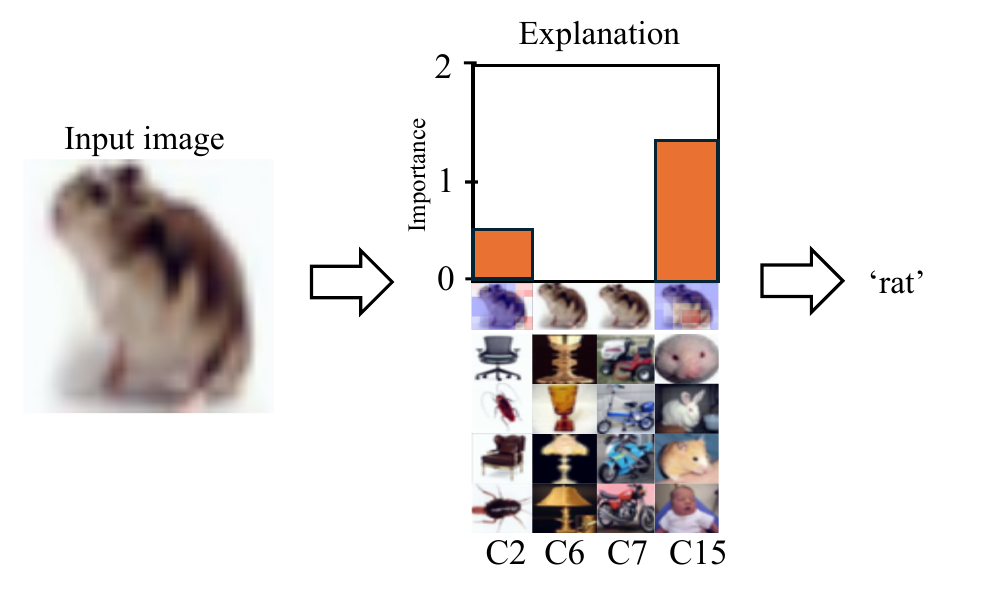}
        \caption{Example belonging to class "rat" extracted form the CIFAR-100 dataset.}
        \label{fig:rat}
    \end{minipage}
\end{figure*}

The final pair of examples, from CIFAR-100, depict a rodent (Fig.\ref{fig:rat}) and a turtle (Fig.\ref{fig:turtle}). The rodent, similar to the previous CIFAR-10 examples, is correctly classified by the model using concept $C2$, which refers to the white background, and $C15$, which corresponds to rat heads. The explanation for the turtle, however, is particularly intriguing. In addition to $C9$, representing reptiles, and $C10$, referring to water/aquatic animals, concept $C4$ (which pertains to trucks) is also activated. While this concept is incorrectly triggered, it is interesting to note that its contribution to the "turtle" class is negative, as it would likely be more relevant to other classes.

\section{Intervention Experiment Results}
\label{app:interventions}

\begin{figure*}[h]
        \centering
        \includegraphics[width=\linewidth]{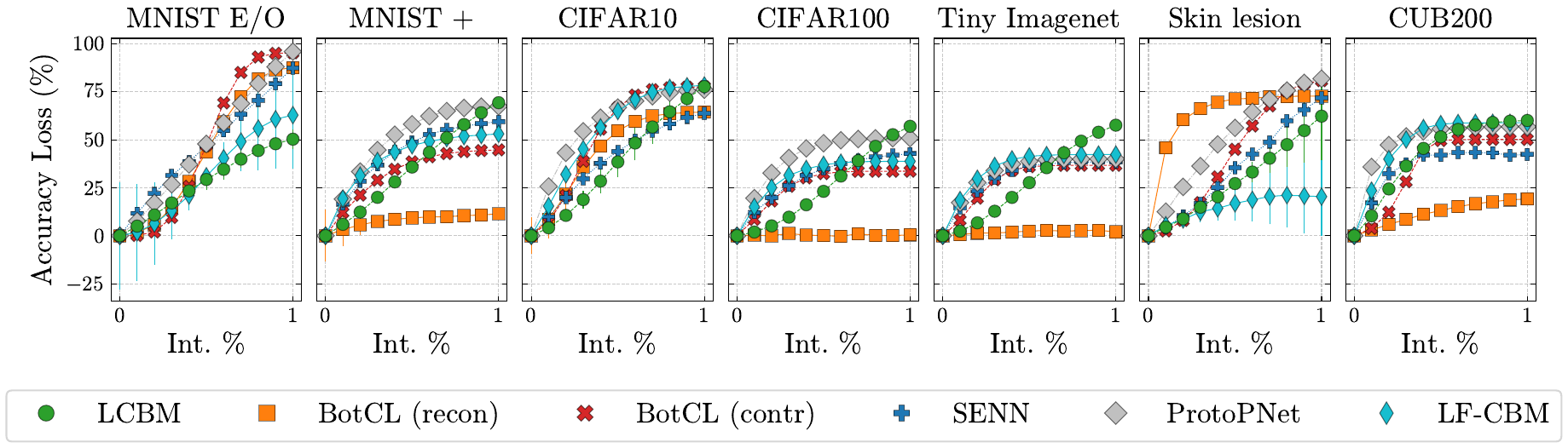}
        \caption{Impact of negative interventions across different datasets. The figure highlights the varying sensitivity of the model, with pronounced effects observed in five datasets while minimal impact is noted in the \textit{MNIST E/O} and \textit{Skin Lesion} datasets.}
        \label{fig:interventions_full}
\end{figure*}

In this appendix, we present the results of the intervention experiment across all datasets. Notably, while our model exhibits minimal sensitivity to negative interventions in the \textit{MNIST E/O} and \textit{Skin Lesion} datasets, it demonstrates a high degree of sensitivity to negative interventions in the remaining five datasets, as illustrated in Fig.~\ref{fig:interventions_full}.

\end{document}